\documentclass[letterpaper, 10 pt, conference]{ieeeconf}  % Comment this line out if you need a4paper

\IEEEoverridecommandlockouts                              % This command is only needed if 
\usepackage{times}
\usepackage{epsfig}
\usepackage{graphicx}
\usepackage{amsmath}
\usepackage{amssymb}
\usepackage{amsfonts}
\usepackage{multirow}
\usepackage{booktabs}    
\usepackage{chngpage}
\usepackage{soul}
\usepackage{hyperref}
\usepackage[flushleft,para]{threeparttable}
\usepackage[table]{xcolor}

\hypersetup{
    colorlinks=true,
    linkcolor=blue,    
    urlcolor=cyan
}

\newcommand{\real}{\mathbb{R}}

\newcommand{\figref}[1]{Fig.~\ref{#1}}
\newcommand{\tabref}[1]{Table~\ref{#1}}
\newcommand{\secref}[1]{Sec.~\ref{#1}}
\newcommand{\eqnref}[1]{Eqn.~\ref{#1}}

\title{\LARGE \bf
Learning Topology from Synthetic Data for \\ Unsupervised Depth Completion
}

\author{Alex Wong$^{1}$, Safa Cicek$^{2}$ and Stefano Soatto$^{1}$% <-this % stops a space
\thanks{This work was supported by ONR N00014-19-1-2229 and ARO W911NF-17-1-0304.}
\thanks{$^{1}$Alex Wong and Stefano Soatto are with Department of Computer Science,
        University of California, Los Angeles. Email:
        {\tt\small alexw@cs.ucla.edu}, {\tt\small soatto@cs.ucla.edu}}
\thanks{$^{2}$Safa Cicek is with the Department of Electrical and Computer Engineering,
        University of California, Los Angeles. Email:
        {\tt\small safacicek@ucla.edu}}%
}

\begin{document}

\maketitle
\thispagestyle{empty}
\pagestyle{empty}

\begin{abstract}
We present a method for inferring dense depth maps from images and sparse depth measurements by leveraging synthetic data to learn the association of sparse point clouds with dense natural shapes, and using the image as evidence to validate the predicted depth map. Our learned prior for natural shapes uses only sparse depth as input, not images, so the method is not affected by the covariate shift when attempting to transfer learned models from synthetic data to real ones. This allows us to use abundant synthetic data with ground truth to learn the most difficult component of the reconstruction process, which is topology estimation, and use the image to refine the prediction based on photometric evidence. Our approach uses fewer parameters than previous methods, yet, achieves the state of the art on both indoor and outdoor benchmark datasets. Code available at: \\ \hyperlink{https://github.com/alexklwong/learning-topology-synthetic-data}{https://github.com/alexklwong/learning-topology-synthetic-data}
\end{abstract}

\section{Introduction}
Images are ``dense'' in the sense of providing a color value (irradiance) at every pixel, but they contain only sparse information about the geometry of the scene, both because (i) the pre-image of a pixel is an arbitrarily large subset of the scene with no single depth value, and (ii) large portions of the image do not allow establishing unique correspondence due to occlusions or the aperture problem. We focus on the second problem (ii), aiming to use higher-level information to impute a depth value at every pixel even when correspondence is not defined (occlusion), or where it yields a continuum of possible depth values (aperture problem).  Where the given images do not provide direct evidence on the geometry of the underlying scene, we have to use priors learned from {\em different} scenes: The fact that walls tend to be flat, surfaces piecewise smooth, objects mostly convex etc. can be evinced from data about scenes other than the one at hand.

The key challenge in this process is to determine the topology of the scene; that is, what point is ``close'' to which in the sense of being part of the same surface or object. Once that is determined,  dense depth estimation is just a matter of piecewise smooth interpolation. Errors in the former cannot be compensated by however clever processing in the latter. So, we focus on {\em learning} scene topology from images, in such a way that can be used to complete the depth map where current images do not provide sufficient evidence.

\begin{figure}[t]
\centering
\includegraphics[width=1.00\linewidth]{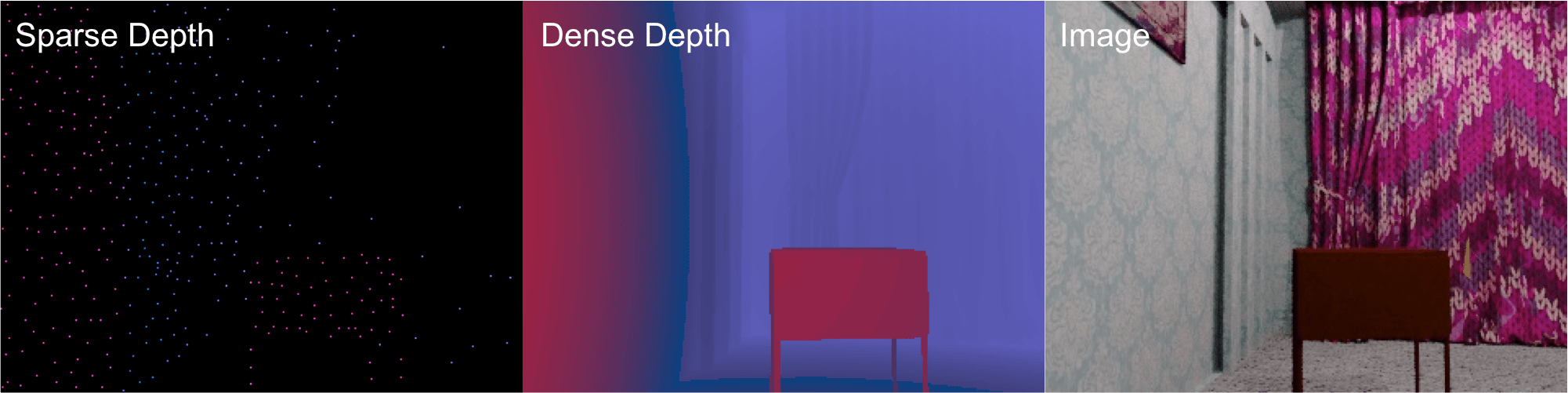}
\caption{\textit{Is it possible learn topology only from sparse points?} There exists an abundance of synthetic data (e.g. SceneNet \cite{mccormac2016scenenet}) with ground-truth depth. We aim to infer the topology of objects from only sparse points (i.e. \textit{without} images), so that we can leverage synthetic data without the need to adapt for the large domain gap between real and synthetic images.}
\label{fig:motivation}
\vspace{-1.3em}
\end{figure}

In some cases, there may be independent mechanisms to associate measurements of light (irradiance at the pixels) to geometric properties of a scene, from tactile perception to controlled illumination or other sensory devices. In a passive setting, absent any side information, it is difficult to obtain ground truth association, so synthetic data is a natural choice but for the covariate shift when transferring models to the real world. We bypass this challenge altogether by learning the association {\em not} from photometry to geometry (images to shapes),  which requires a high-level understanding of the semantics of objects, but from sparse geometry (point cloud) to topology (connectivity and dense geometry), using abundant synthetic data, without having to face concerns about covariate shift and domain adaptation.

Recent approaches to depth completion that are amenable to utilize our model include \cite{wong2020unsupervised,yang2019dense}. \cite{yang2019dense} utilizes both image and depth from synthetic data and is plagued by the domain gap when transferring to real data. \cite{wong2020unsupervised} uses a piecewise planar ``scaffolding'' of the scene to transfer the supervisory signal from sparse points to their neighbors. However, the piecewise planar assumption is too coarse and initial errors can have catastrophic consequences.  We learn the topology of the objects only from sparse points (i.e. no RGB images). Of course, the learned topology is only a ``guess,''  or prior, which needs to be reconciled with the images, but we posit that it is better than generic priors like smoothness or proximity, as it enables drawing from properties of the distribution of natural shapes. Accordingly, we first learn to predict an \textit{approximate topology} from sparse points with a lightweight network that we call ScaffNet. In a second stage, our fusion network (FusionNet) leverages photometric evidence from real images to correct the prediction.

More specifically, our \textbf{contributions} include (i) repurposing Spatial Pyramid Pooling (SPP) \cite{he2015spatial} to densify the sparse depth measurements while balancing the trade-off between sparsity (pooling with small kernels) and levels of detail (large kernels); (ii) we train an SPP-augmented, light-weight network (ScaffNet) on synthetic data to learn connectivity and dense geometry from sparse inputs and demonstrate that the shapes learned can generalize well to datasets with different scene geometry (i.e. from synthetic scenes of randomly arranged household rooms in SceneNet \cite{mccormac2016scenenet} to real scenes of laboratories, classrooms and gardens in VOID \cite{wong2020unsupervised}); (iii) we propose to learn additive and multiplicative residuals with FusionNet to alleviate the network from the burden of having to re-learn depth; (iv) we treat the topology learned by ScaffNet as a prior and design an adaptive loss function that selectively regularizes the predictions of FusionNet by conditioning on the fitness of the each model to data.

\section{Related Work}
\textbf{Supervised depth completion}  methods learn a direct map from image and sparse depth to dense depth by minimizing the difference to ground-truth. \cite{chodosh18deep} cast depth completion as compressive sensing by learning a dictionary and \cite{dimitrievski2018learning} morphological operators. Recent innovations include network operations \cite{eldesokey2018propagating,huang2019hms} and architectures \cite{chen2019learning,ma2019self,uhrig2017sparsity,yang2019dense} for processing sparse depth. \cite{ma2019self} handled sparse depth and images separately and fused them after a single convolution (early fusion), while \cite{jaritz2018sparse,yang2019dense} used two separate encoders (late fusion); \cite{chen2019learning} used a 2D-3D fusion network. To propagate sparse depth  through the network, \cite{eldesokey2018propagating} used normalized convolutions with a binary map while \cite{huang2019hms} performed joint concatenation and convolution to upsample the sparse depth. Additionally, \cite{qu2021bayesian,qu2020depth,van2019sparse} learned confidence maps and \cite{qiu2019deeplidar,xu2019depth,zhang2018deep} exploited surface normals for guidance. Like us, \cite{merrill2021robust,sartipi2020deep,zuo2021codevio} proposed light-weight networks that can be deployed onto SLAM/VIO systems.

Training these methods requires per-pixel ground-truth, often unavailable or prohibitively expensive. We instead learn to infer topology from synthetic data with ground truth and abundant un-annotated real data.

\textbf{Unsupervised depth completion}  assumes additional (stereo, temporally consecutive frames) data available during training. Both stereo \cite{shivakumar2019dfusenet,yang2019dense} and monocular \cite{ma2019self,wong2020targeted,wong2021adaptive,wong2020unsupervised} paradigms learn dense depth from an image and sparse depth measurements by minimizing the photometric error between the input image and its reconstruction from other views along with the difference between prediction and sparse depth input (sparse depth reconstruction). \cite{ma2019self} used Perspective-n-Point  \cite{lepetit2009epnp} and RANSAC \cite{fischler1981random} to align consecutive frames and \cite{wong2019bilateral,wong2021adaptive} proposed an adaptive weighting framework. \cite{yang2019dense} also used synthetic data but require image, sparse and ground-truth depth. They do not address the domain gap between the additional dataset and the target dataset; hence, their learned prior may at times hurt performance. Unlike \cite{yang2019dense}, we seek to learn topology from sparse points from a synthetic dataset and do not require images, thus bypassing the domain gap. \cite{ma2019self,shivakumar2019dfusenet,yang2019dense} learn end-to-end without supervision, with sparse depth input. However, convolutions are ineffective in processing sparse input because most of the receptive fields are not activated in the early layers. Instead, we leverage spatial pyramid pooling \cite{he2015spatial} to increase the receptive field and ``densify'' the sparse input before feeding it into a topology estimation network.
%Unlike \cite{ma2019self,shivakumar2019dfusenet,yang2019dense}, 
\cite{wong2020unsupervised} used a two-stage approach to first approximate the mesh and later fuse with image information. The ``scaffolding'' is prone to errors in regions that lack depth input or contain complex structures.  %Moreover, \cite{wong2020unsupervised} only used it as an input and does not leverage it in the output or loss function. 
Our approach is also two-staged, but we seek to learn topology from synthetic data. To alleviate the fusion network from having to re-learn the approximate geometry, we learn the residual from the image to refine the approximation using a fusion network. Lastly, we also regularize our predictions using the approximated topology conditioned on the fitness of the model to data. 

\textbf{Domain adaptation for depth prediction} \cite{nath2018adadepth} applied ordinary domain adaptation to single-image depth prediction. We do not attempt to reduce the domain gap between synthetic and real images, but leverage the sparse depth to learn dense topology from synthetic shapes. We use RGB images only as evidence to refine the estimates.  \cite{atapour2019generative} leveraged synthetic data for in-painting depth. Our problem is more challenging as the sparse points cover $\approx$5\% of the image space  \cite{uhrig2017sparsity} and as few as 0.5\% indoors \cite{wong2020unsupervised}. % \cite{nath2018adadepth,atapour2018real,zhao2019geometry}

\begin{figure}[t]
\centering
\includegraphics[width=1.00\linewidth]{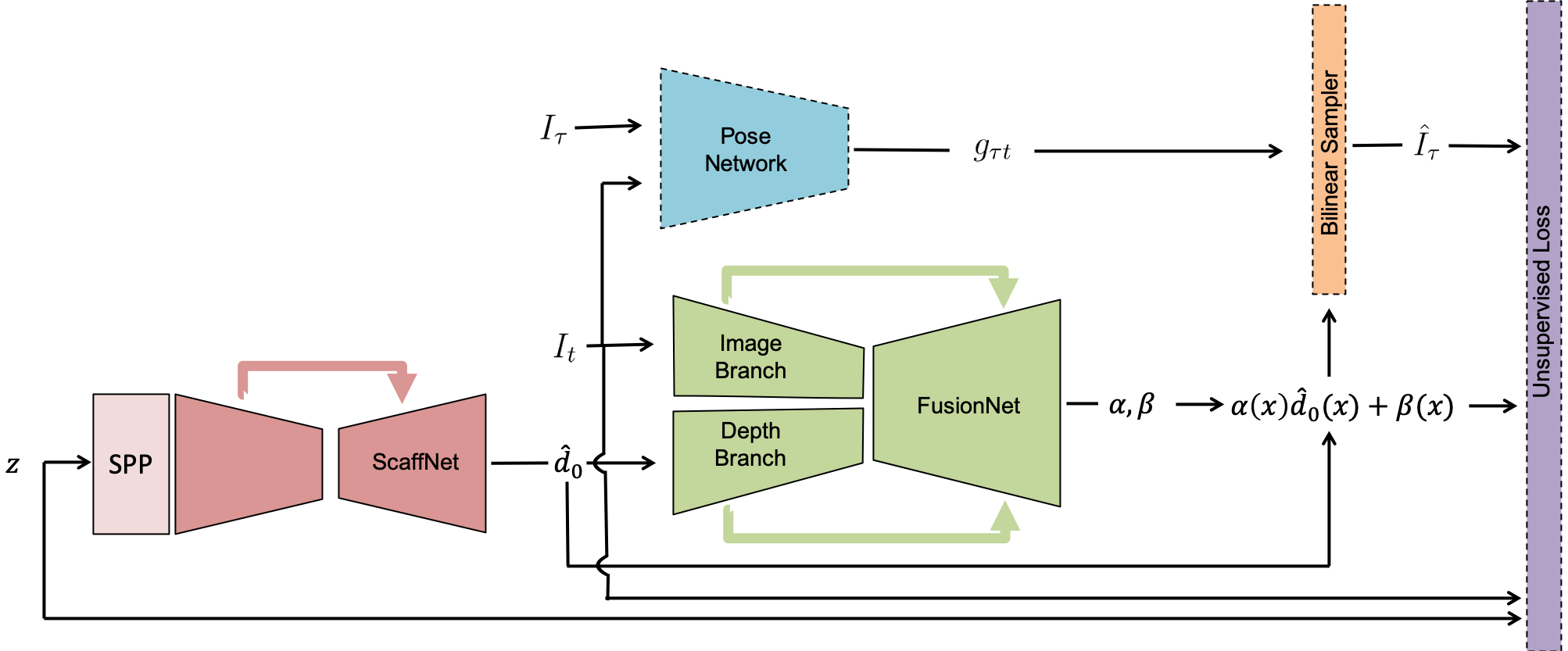}
\caption{\textit{Overview.} Sparse points ($z$) are first densified by SPP (see \figref{fig:spatial_pyramid_pooling}), and fed to ScaffNet (trained with synthetic data) to produce an approximate topology $\hat{d}_{0}$. FusionNet, trained with an unsupervised loss (\eqnref{eqn:objective_function}), refines $\hat{d}_{0}$ with $\alpha$ and $\beta$ by fusing the $\hat{d}_{0}$ with the information from the image $I_t$ to produce the final prediction $\hat{d}(x)=\alpha(x) \hat{d}_0(x) + \beta(x)$ for $x \in \Omega$. Only ScaffNet (red) and FusionNet (green) are required for inference.}
\label{fig:overview}
\vspace{-1.3em}
\end{figure}

\section{Method Formulation}
\label{sec:method_formulation}
Our goal is to recover a 3D scene from a real RGB image $I_t : \Omega \subset \real^2 \mapsto \real^3_+$ and the associated set of sparse depth measurements $z : \Omega_{z} \subset \Omega \mapsto \real_+$, without access to ground-truth depth annotations. We follow the unsupervised monocular training paradigm \cite{ma2019self,wong2020unsupervised} and assume there exists temporally adjacent frames, $I_\tau$ for $\tau \in T \doteq \{t-1, t+1\}$ denoting the previous and the next time stamp relative to $I_t$, available during training. Additionally, we assume there also exists a synthetic dataset, containing sparse depth $z : \Omega_{z} \mapsto \real_+$ and associated ground-truth dense depth $d_{gt} : \Omega  \mapsto \real_+$, available. 

Our approach is separated into two stages: (i) we train a topology estimator $f_\omega(z)$ with a synthetic dataset $\mathcal{D}$ to approximate the scene $\hat{d}_{0} := f_\omega(z)$. By exploiting the statistics of large synthetic datasets (where one can obtain ground-truth depth for free), our topology estimator learns to extract patterns of connectivity between sparse points to model scene structures too complex for hand-crafted priors (e.g. nearest-neighbor). However, as the topology is only informed by the sparse points, one cannot hope to recover regions with very few (or no) points with high precision. This is where the image comes back into the picture; (ii) we refine the initial estimate $\hat{d}_{0}$ by incorporating information from the image belonging to the target (real) domain for which we \textit{do not} have any ground-truth depth. We propose to learn a multiplicative scale $\alpha_{\theta}(I_t, z, \hat{d}_{0})$ and an additive residual $\beta_{\theta}(I_{t}, z, \hat{d}_{0})$ around $\hat{d}_{0}$ where $[\alpha_{\theta}(I_{t}, z, \hat{d}_{0}), \beta_{\theta}(I_{t}, z, \hat d_{0})] = f_{\theta}(I_t, z, \hat d_{0})$\footnote{More formally, $\alpha$ and $\beta$ should be written as a function of position $x \in \Omega$, $\alpha_{\theta}(I_{t}(x), z(x), \hat d_{0}(x))$. For simplicity, we will refer to them as $\alpha(x)$ and $\beta(x)$ without the parameters and inputs.}. In other words, our network $f_{\theta}$ fuses the image information ($I_{t}$) with target sparse points ($z$) and initial estimate ($\hat{d}_{0}$) to produce the final dense depth $\hat{d}$ (see \figref{fig:overview}). Hence, the name FusionNet. By learning $\alpha(x)$ and $\beta(x)$ around $\hat{d}_0(x)$ for $x \in \Omega$, instead of directly mapping from the initial estimate and image to the final prediction, we alleviate the network from having to re-learn depth from scratch. 

\subsection{Learning Topology using Synthetic Data}
\label{sec:learning_topology}
Can we learn to infer the dense topology of the scene given \textit{only} sparse points? This is an ill-posed problem as there exists infinitely many possible scenes compatible with the missing depth measurements. We propose to learn a topology estimator (ScaffNet) to produce dense depth from sparse depth measurements \textit{without} the use of an RGB image. To accomplish this, we leverage synthetic datasets with accurate dense depth to capture the patterns of complex geometry, present in everyday objects, that hand-crafted priors (e.g. nearest neighbor, piece-wise smoothness \cite{wong2020unsupervised}) cannot.

For this, we train a small encoder-decoder network $f_{\omega}(\cdot)$, comprised of only $\approx$1.4M parameters, by minimizing the normalized $L1$ difference between the estimate $\hat{d}_{0}$ and the ground-truth dense depth $d_{gt}$ for each sample:  
\begin{equation}
    l_{0} = \frac{1}{|\Omega|}\sum_{x \in \Omega} |\frac{f_\omega(z(x)) - d_{gt}(x)}{d_{gt}(x)}|
\end{equation}
 to learn a mapping from sparse points $z$ to dense topology $\hat{d}_{0} = f_\omega(z(x))$. We note that it is impossible for $f_{\omega}(\cdot)$ to recover regions of the scene that lack sparse depth measurements. Hence, $\hat{d}_{0}$ serves as an initial estimate of the scene. Therefore, we propose to refine $\hat{d}_{0}$ using information from the image (see \secref{sec:learning_to_refine}).

\begin{figure}[t]
    \centering
    \includegraphics[width=1.00\linewidth]{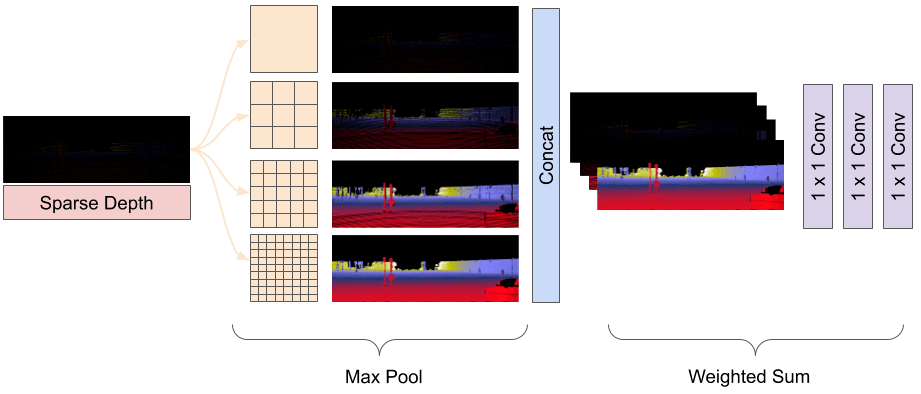}
    \caption{\textit{Spatial Pyramid Pooling (SPP) for depth completion}. The SPP module balances details versus density for sparse depth inputs max-pooled at different scales. When pooled with small kernels, the details of the sparse inputs are preserved, but the subsequent layers will produce very few non-zero activations. When pooled with large kernels, inputs are densified, but details are lost. By weighting the output of the pooling with several stacked $1 \times 1$ convolutions, the network learns to optimize for this trade-off.}
\label{fig:spatial_pyramid_pooling}
\vspace{-1.3em}
\end{figure}

\textbf{Spatial Pyramid Pooling (SPP).} Since most of the input values for sparse depth measurements are zeroes, the activations of early convolutional layers tend to be zeroes as well. To process the sparse input, we augment our network ($f_{\omega}(\cdot)$) with an SPP module (see \figref{fig:spatial_pyramid_pooling}) where each layer in SPP is a max-pool of different kernel sizes. For max-pool layers with large kernel sizes, the sparse input is densified, leading to more neurons being activated in the subsequent layers. However, the finer details (e.g. small or thin objects close to the camera) are corrupted. For max-pool layers with small kernel sizes, details of the sparse input are preserved, but as a result, fewer neurons are activated. Therefore, we weight the output of the max-pool layers, with different kernel-sizes, using several stacked $1 \times 1$ convolutional layers such that the network can optimize for this trade-off. We note that SPP also increases the receptive field of the network. The output of SPP is fed into our encoder-decoder network and the weights belonging to the $1 \times 1$ convolutions are jointly optimized. We demonstrate the effectiveness of our SPP module in an ablation study in \tabref{tab:kitti_validation_set_ablation} of \secref{sec:experiments_results}. To validate our claim that the representation obtained without SPP is much sparser than that obtained with SPP, we provide additional discussion, ablation studies, and visualizations of features extracted with and without SPP in Sec. I of Supp. Mat. We also discuss differences between our variant of SPP and those employed in classification \cite{chang2018pyramid} and stereo \cite{he2015spatial} (both operating on \textit{already dense} inputs) in Sec. I-E of Supp. Mat. 

\subsection{Learning to Refine (Bringing the Image Back)}
\label{sec:learning_to_refine}
ScaffNet, our topology estimator, $f_\omega(\cdot)$ learns to predict the coarse scene structure $\hat{d}_{0}$ from sparse points $z$ where available. However, regions where sparse points are unavailable (due to scan pattern or range of sensor), predictions of $f_\omega(\cdot)$ can be misleading. This is where we bring the image back. We want to learn a function $f_{\theta}(I_{t}, z, \hat d_0)$ that produces the scale $\alpha(x)$ and the residual $\beta(x)$ for every element in the image domain, $x \in \Omega$. $\alpha(x)$ and $\beta(x)$ refine each $\hat{d}_0(x)$ based on the image to produce the final depth prediction:
\begin{equation}
    \hat{d}(x) = \alpha(x) \hat{d}_0(x) + \beta(x)
\end{equation} 
where $\alpha$ is encouraged to be close to 1 and $\beta$ to be close to 0 (see \eqnref{eqn:topology_prior_loss}). To learn $\alpha(x)$ and $\beta(x)$, we leverage the geometric relations between the image $I_{t}$ and its temporally adjacent frames $I_{\tau}$, which provide a set of constraints on depth values, up to an unknown scale. These constraints are realized by (i) reconstructing $I_{t}$ with $I_{\tau}$ to construct a photometric consistency loss (\eqnref{eqn:photometric_consistency_loss}). To ground the depth values to metric scale, we (ii) reconstruct the sparse depth measurements $z$ (where available) with $\hat{d}$ (\eqnref{eqn:sparse_depth_consistency_loss}). As depth completion is ill-posed, we assume (iii) generic (not informed by data) local smoothness and connectivity (\eqnref{eqn:local_smoothness_loss}) on the surfaces populating the scene. Lastly, to leverage the prior learned from synthetic data (side information), we propose to (iv) \textit{selectively} regularize $\hat{d}$ using $\hat{d}_{0}$ conditioned on the fitness of model to the data (\eqnref{eqn:topology_prior_loss}). Our unsupervised objective function is the linear combination of four terms:
\begin{equation}
    \label{eqn:objective_function}
    \mathcal{L} = w_{ph}l_{ph}+w_{sz}l_{sz}+w_{sm}l_{sm}+w_{pz}l_{pz}
\end{equation}
where $l_{ph}$ denotes photometric consistency, $l_{sz}$ sparse depth consistency, $l_{sm}$ local smoothness and $l_{pz}$ the topology prior. Each term is weighted by their associated $w$ (see \secref{sec:implementation_details}).

\textbf{Photometric Consistency.}
We leverage epipolar constraints as a supervisory signal by reconstructing $I_{t}$ from $I_{\tau}$ for $\tau \in T \doteq \{t-1, t+1\}$ to yield:
\begin{equation}
\label{eqn:image_reconstruction}
    \hat{I}_\tau(x, \hat d) = I_{\tau} \big( \pi  g_{\tau t} K^{-1} \bar{x} \hat d(x) \big)
\end{equation}
where $\bar{x} = [x^\top \ 1]^\top$ are the homogeneous coordinates of $x \in \Omega$, $g_{\tau t} \in SE(3)$ is the relative pose of the camera from time $t$ to $\tau$, $K$ denotes the camera intrinsics, and $\pi$ refers to the perspective projection. To realize the geometric constraints as a loss, we minimize the average photometric reprojection error measured by a combination of $L1$ penalty and \texttt{SSIM} \cite{wang2004image}, a perceptual metric, to penalize dissimilarities:
\begin{equation}
\begin{aligned}
  	l_{ph} = \frac{1}{|\Omega|} 
  	    \sum_{\tau\in T} \sum_{x \in \Omega}  
  	        &w_{co}| \hat{I}_{\tau}(x, \hat d)-I_{t}(x)| + \\ 
  	        &w_{st}\big(1 - \texttt{SSIM}(\hat{I}_{\tau}(x, \hat d), I_t(x))\big)
\label{eqn:photometric_consistency_loss}
\end{aligned}
\end{equation}
$w_{co}$ and $w_{st}$ are weights for each term and will be discussed in \secref{sec:implementation_details}. Note that we also jointly learn pose $g_{\tau t}$ as a by product of minimizing \eqnref{eqn:photometric_consistency_loss}. 

\textbf{Sparse Depth Consistency.}
Photometric reconstruction recovers the scene structure up to a scale; to ground the scene to \textit{metric} scale, we minimize the $L1$ difference between our predictions $\hat{d}$ and the sparse depth measurements over the sparse depth domain ($\Omega_{z}$):
\begin{equation}
\label{eqn:sparse_depth_consistency_loss}
  	l_{sz} = \frac{1}{|\Omega_z|} 
  	    \sum_{x \in \Omega_z} 
  	        | \hat{d}(x) - z(x)|. 
\end{equation}

\textbf{Local Smoothness.}
While sparse depth measurements removes ambiguity, there still exists infinitely many scenes compatible with the image in regions \textit{not} in the sparse depth domain ($\Omega \backslash \Omega_{z}$). Hence, we assume local smoothness and connectivity over $\hat{d}$ with an $L1$ penalty on the gradients in the $x-$ ($\partial_{X}$) and $y-$ ($\partial_{X}$) directions. To allow discontinuities along object boundaries, we weight each term using its respective image gradients, $\lambda_{X} = e^{-|\partial_{X}I_{t}(x)|}$ and $\lambda_{Y} = e^{-|\partial_{Y}I_{t}(x)|}$, where strong gradients produce lower weight:
\begin{equation}
\label{eqn:local_smoothness_loss}
  	l_{sm} = \frac{1}{|\Omega|}
  	    \sum_{x \in \Omega} 
      	    \lambda_{X}(x)|\partial_{X}\hat{d}(x)|+
      	    \lambda_{Y}(x)|\partial_{Y}\hat{d}(x)|.
\end{equation}

\textbf{Topology Prior.}
Learning depth from motion is essentially a correspondence problem (2D search space over the image domain). Rather than exploring the entire solution or hypothesis space, we look to leverage the initial topology $\hat{d}_0$ predicted by $f_\omega(\cdot)$ as a prior to limit the scope or to bias our predictions towards the set of hypotheses compatible with what we have learned from a synthetic dataset, $\mathcal{D}$. However, the quality of $\hat{d}_0$ degrades in regions with very few or no sparse depth measurements. Hence, one should $\textit{selectively}$ regularize based on the compatibility between the prior $\hat{d}_0$ and the image $I_{t}$. To elaborate, we should bias our predictions $\hat{d}$ towards the prior $\hat{d}_0$ only if $\hat{d}_0$ is more compatible with the image than $\hat{d}$. Following this intuition, we present a simple, yet effective, per-pixel regularizer conditioned on the fitness of $\hat{d}_0$ and $\hat{d}$ to the image as measured by the discrepancies between their respective reconstructions (\eqnref{eqn:image_reconstruction}), $\hat{I}_{\tau}(x, \hat{d})$ and $\hat{I}_{\tau}(x, \hat{d}_0)$,  and the image $I_{t}(x)$.

To optimize for this trade-off, we construct $W(x)$, an indicator function that is $1$ if the photometric discrepancy $\delta = |I_{t}(x) - \hat{I}_{\tau}(x, \hat{d})|$ is greater than $\delta_0 = |I_{t}(x) - \hat{I}_{\tau}(x, \hat{d}_{0})|$ and $0$ otherwise, for every $x \in \Omega$,
\begin{equation}
    W(x) = \begin{cases} 
        1 &\mbox{if } \delta > \delta_0 \\
        0 & \mbox{otherwise}.   
    \end{cases}    
\end{equation}
We then use $W(x)$ as a pixel-wise mask to selectively impose the topology prior $\hat{d}_{0}(x)$ on the predictions $\hat{d}(x)$:
\begin{equation}
\label{eqn:topology_prior_loss}
    l_{tp} =  \frac{1}{\displaystyle \sum_{x \in \Omega}{W(x)}}
        \sum_{x \in \Omega}
            W(x) |\hat{d}(x) - \hat{d}_{0}(x)|.
\end{equation}
Note that \eqnref{eqn:topology_prior_loss} is flexible and encourages $\alpha(x)$ to be close to 1 and $\beta(x)$ to 0 \textit{only} at locations where the initial topology prediction $\hat{d}_{0}$ is already a good fit for the image. The network is free to modify $\hat{d}_{0}$ where the topology is incorrect.

\begin{table}[t]
\caption{Error metrics}
\centering
\setlength\tabcolsep{22pt}
\begin{tabular}{l l}
\midrule
    Metric & Definition \\ \midrule
    MAE &$\frac{1}{|\Omega|} \sum_{x\in\Omega} |\hat d(x) - d_{gt}(x)|$ \\
    RMSE & $\big(\frac{1}{|\Omega|}\sum_{x\in\Omega}|\hat d(x) - d_{gt}(x)|^2 \big)^{1/2}$ \\
    iMAE & $\frac{1}{|\Omega|} \sum_{x\in\Omega} |1/ \hat d(x) - 1/d_{gt}(x)|$ \\
    iRMSE& $\big(\frac{1}{|\Omega|}\sum_{x\in\Omega}|1 / \hat d(x) - 1/d_{gt}(x)|^2\big)^{1/2}$ \\ \midrule
\end{tabular}
\begin{tablenotes}
    Evaluation metrics for KITTI and VOID. $d_{gt}$ denotes the ground truth.
\end{tablenotes}
\label{tab:error_metrics}
\vspace{-2.0em}
\end{table}

\begin{figure*}[th]
    \centering
    \includegraphics[width=1\linewidth, height=0.27\linewidth]{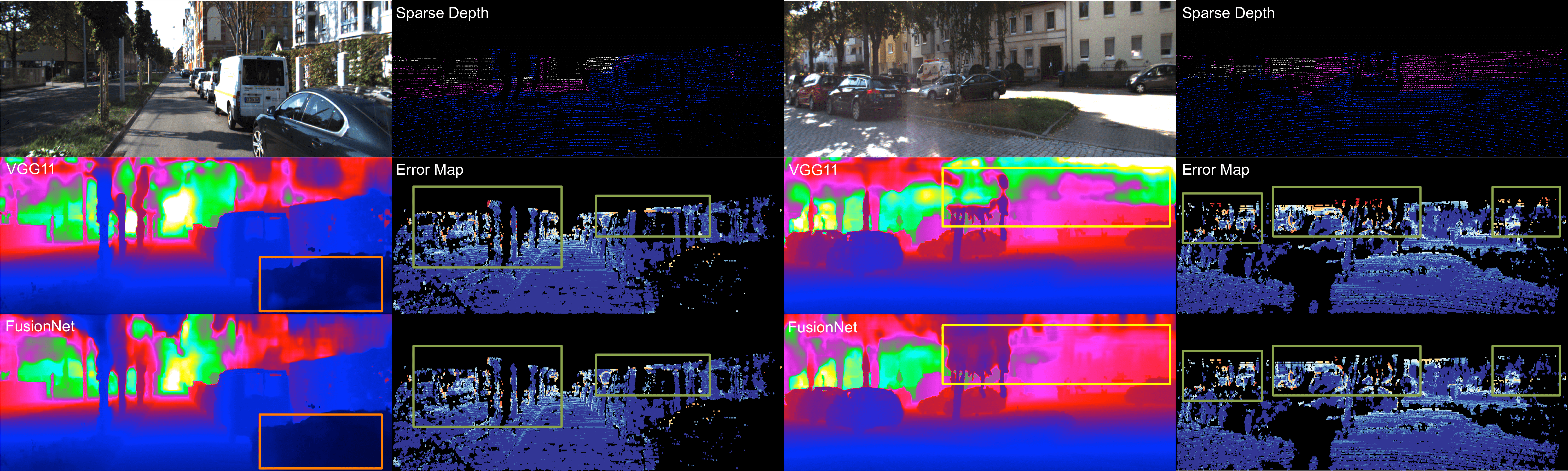}
    \caption{\textit{KITTI depth completion testing set.}  Visualizations are taken directly from KITTI online benchmark. Our method better captures smooth surfaces (car, highlighted in orange) and complex objects (trees, highlighted in yellow). We also perform better in recovering buildings (also in yellow). The overall improvement is apparent in the error map. Many red regions (high error) in \cite{wong2020unsupervised} are marked blue (low error) in our results.}
    \label{fig:kitti_testing_results}
    \vspace{-1.5em}
\end{figure*}

\section{Datasets}
\label{sec:datasets}
\textbf{SceneNet} \cite{mccormac2016scenenet} consists of 5 million RGB image and depth map pairs of size $320 \times 240$ rendered from indoor trajectories of randomly arranged rooms. Out of 17 splits provided, we only choose one due to computational restrictions. In a single split, there are 1000 subsequences, each containing 300 images of the same scene, recorded over a trajectory. We constructed sparse points for each ground-truth depth sample by running Harris corner detector \cite{harris1988combined} over the RGB images. The resulting points are subsampled using k-means to produce the final 375 corners which corresponds to 0.49\% of all the pixels. We train ScaffNet on SceneNet for indoor target dataset (VOID \cite{wong2020unsupervised}). Despite SceneNet (indoor, household rooms) having different scene structures from VOID (indoor and outdoor, laboratories and gardens), ScaffNet trained on SceneNet generalizes to VOID. 
% 17*1000*300
% 100*375/(240*320)=0.49

\textbf{Virtual KITTI (VKITTI)} \cite{gaidon2016virtual} consists of 35 synthetic videos (5 cloned from the KITTI \cite{uhrig2017sparsity}, each with 7 variations in weather, lighting or camera angle) for a total of $\approx$17K $1242 \times 375$ frames.  \cite{gaidon2016virtual} use the commercial computer graphics engine Unity to create virtual worlds that similar to scenes in KITTI. However, there is still a large domain gap between RGB images of both domains. To bypass the domain gap in photometric variations, we only use the dense depth maps of VKITTI. To acquire the sparse points, we imitate the sparse depth measurement of KITTI (produced by a lidar) so that the marginal distributions of sparse points are close across domains. We use VKITTI to train the topology estimator for outdoor target dataset (KITTI). 

\textbf{KITTI.} We evaluate our approach on the KITTI depth completion benchmark \cite{uhrig2017sparsity}. The dataset provides $\approx$80,000 raw image frames and associated sparse depth maps. The sparse depth maps are the raw output from the Velodyne lidar sensor, each with a density of $\approx$5\%. The ground-truth depth map is created by accumulating the neighbouring 11 raw lidar scans, with dense depth corresponding to the bottom $30\%$ of the images. We use the official 1,000 samples for validation and test on 1,000 designated samples, which we submit to the KITTI online benchmark for evaluation. 

\textbf{VOID} \cite{wong2020unsupervised} provides synchronized RGB image frames and sparse depth maps of $\approx640 \times 480$  resolution of indoor (laboratories, classrooms) and outdoor (gardens) scenes. $\approx 1500$ sparse depth  points (covering $\approx0.5\%$ of the image) are the set of features tracked by XIVO \cite{fei2019geo}, a VIO system. The ground-truth depth maps are dense and are acquired by active stereo. The entire dataset contains 56 sequences with challenging motion. Of the 56 sequences, 48 sequences ($\approx 40,000$) are designated for training and 8 for testing. The testing set contains $800$ frames. We follow the evaluation protocol of \cite{wong2020unsupervised} and cap the depths between 0.2 and 5 meters.

\begin{table}[t]
    \caption{Results on the KITTI validation set}
    \setlength\tabcolsep{6pt}
    \centering
    \begin{tabular}{l c c c c c}
        \midrule 
        Method & \# Param & MAE & RMSE & iMAE & iRMSE \\ 
        \midrule 
        Scaffolding \cite{wong2020unsupervised} 
        & 0 & 443.57 & 1990.68 & 1.72 & 6.43 \\ 
        \midrule
        Our ScaffNet 
        & $\approx$1.4M & 318.41 & 1425.53 & 1.39 & 5.01 \\ 
        \midrule
        Ma \cite{ma2019self} 
        & $\approx$27.8M & 358.92 & 1384.85 & 1.60 & 4.32 \\ 
        \midrule
        Yang \cite{yang2019dense}
        & $\approx$18.8M & 347.17 & 1310.03  & n/a & n/a \\ 
        \midrule
        VGG8 \cite{wong2020unsupervised}   
        & $\approx$6.4M & 308.81 & 1230.85 & 1.29 & 3.84 \\ 
        \midrule
        VGG11 \cite{wong2020unsupervised} 
        & $\approx$9.7M & 305.06 & 1239.06 & 1.21 & 3.71 \\ 
        \midrule
        Our FusionNet 
        & $\approx$7.8M & \textbf{286.35} & \textbf{1182.81} & \textbf{1.18} & \textbf{3.55} \\
        \midrule
    \end{tabular}
    \begin{tablenotes}
        Results of \cite{ma2019self,wong2020unsupervised,yang2019dense} are directly taken from their papers. Our method (FusionNet) performs the best across all metrics for the KITTI validation set while using fewer parameters than the state of the art (VGG11 \cite{wong2020unsupervised}). Our topology estimator (ScaffNet) alone outperforms \cite{ma2019self,yang2019dense} on MAE and \cite{ma2019self} on iMAE (\cite{yang2019dense} did not report their iMAE on the validation set). Note that Scaffolding and Our ScaffNet are using sparse-depth only and do not use RGB images. ScaffNet is trained on synthetic data and evaluated on real data.
    \end{tablenotes}
\label{tab:kitti_validation_set_results}
\vspace{-1.5em}
\end{table}

\section{Implementation Details}
\label{sec:implementation_details}
To train our system, we: (i) train ScaffNet on synthetic data, (ii) freeze ScaffNet weights, (iii) train FusionNet with frozen ScaffNet on real data. Training ScaffNet on VKITTI \cite{gaidon2016virtual} took $\approx$12 hours (30 epochs) while training FusionNet on KITTI \cite{uhrig2017sparsity} requires  $\approx$27 hours (30 epochs) on an Nvidia GTX 1080Ti. Training ScaffNet on SceneNet \cite{mccormac2016scenenet} requires $\approx$6 hours (10 epochs). FusionNet also requires $\approx$6 hours (10 epochs) for VOID \cite{wong2020unsupervised}. End to end inference takes $\approx$32 ms per image. We used Adam \cite{kingma2015adam} with $\beta_1=0.9$ and $\beta_2=0.999$ to optimize our network with a base learning rates of $1.5 \times 10^{-4}$ for VKITTI and KITTI and $5 \times 10^{-5}$ for SceneNet and VOID. We decrease the learning rate by half after 18 epochs for VKITTI and KITTI and 6 epochs for SceneNet and VOID, and again after 24 epochs and 8 epochs, respectively. We train our network with a batch size of 8 using $768 \times 320$ crops for VKITTI and KITTI and $640 \times 480$ for SceneNet and VOID. To replicate our results on KITTI, we set the weights for each term in our loss function as: $w_{ph}=1.00$, $w_{co}=0.20$, $w_{st}=0.40$, $w_{sz}=0.10$, $w_{sm}=0.01$ and $w_{tp}=0.10$. For VOID, we increased $w_{sz}$ to $1.00$ and $w_{sm}$ to $0.40$. We only apply $l_{tp}$ after 60K steps (for 271K total steps) for KITTI and 20K steps (for 51K total steps) for VOID to allow pose to stabilize. We perform horizontal shifts as data augmentation on KITTI and VKITTI. We do not use any data augmentation for SceneNet and VOID.  

\begin{table}[t]
    \caption{Results on the KITTI depth completion benchmark}
    \setlength\tabcolsep{6pt}
    \centering
    \begin{tabular}{l c c c c c}
        \midrule
        Method & \# Param & MAE & RMSE & iMAE & iRMSE \\ 
        \midrule
        Schneider \cite{schneider2016semantically}   
        & n/a & 605.47 & 2312.57 & 2.05 & 7.38 \\ 
        \midrule
        Ma \cite{ma2019self}
        & $\approx$27.8M & 350.32 & 1299.85 & 1.57 & 4.07 \\ 
        \midrule
        Yang \cite{yang2019dense}
        & $\approx$18.8M & 343.46 & 1263.19 & 1.32 & 3.58 \\ 
        \midrule
        Shivakumar \cite{shivakumar2019dfusenet}
        & n/a & 429.93 & 1206.66 & 1.79 & 3.62 \\ 
        \midrule
        VGG8 \cite{wong2020unsupervised}  
        & $\approx$6.4M & 304.57 & 1164.58 & 1.28 & 3.66 \\ 
        \midrule
        VGG11 \cite{wong2020unsupervised}  
        & $\approx$9.7M & 299.41 & 1169.97 & 1.20 & 3.56 \\ 
        \midrule
        Our FusionNet
        & $\approx$7.8M & \textbf{280.76} & \textbf{1121.93} & \textbf{1.15}  & \textbf{3.30} \\
        \midrule
    \end{tabular}
    \begin{tablenotes}
        Results are directly taken from the KITTI online benchmark. Key comparisons: (i) \cite{yang2019dense} uses both synthetic images and depth maps to train their model and is plagued by the domain gap between real and synthetic images. We bypass the need to adapt to the covariate shift by learning topology from only sparse depth. (ii) \cite{wong2020unsupervised} uses hand-crafted scaffolding and learns depth from scratch. Instead, we propose to exploit the distribution of natural shapes from synthetic datasets and learn multiplicative and additive residuals to alleviate the network from needed to re-learn depth. Our approach beats all competing methods across all metrics and achieves the state of the art on the unsupervised depth completion task.
    \end{tablenotes}
\label{tab:kitti_testing_set_results}
\vspace{-1.5em}
\end{table}

\begin{figure*}[t]
    \centering
    \includegraphics[width=0.80\linewidth]{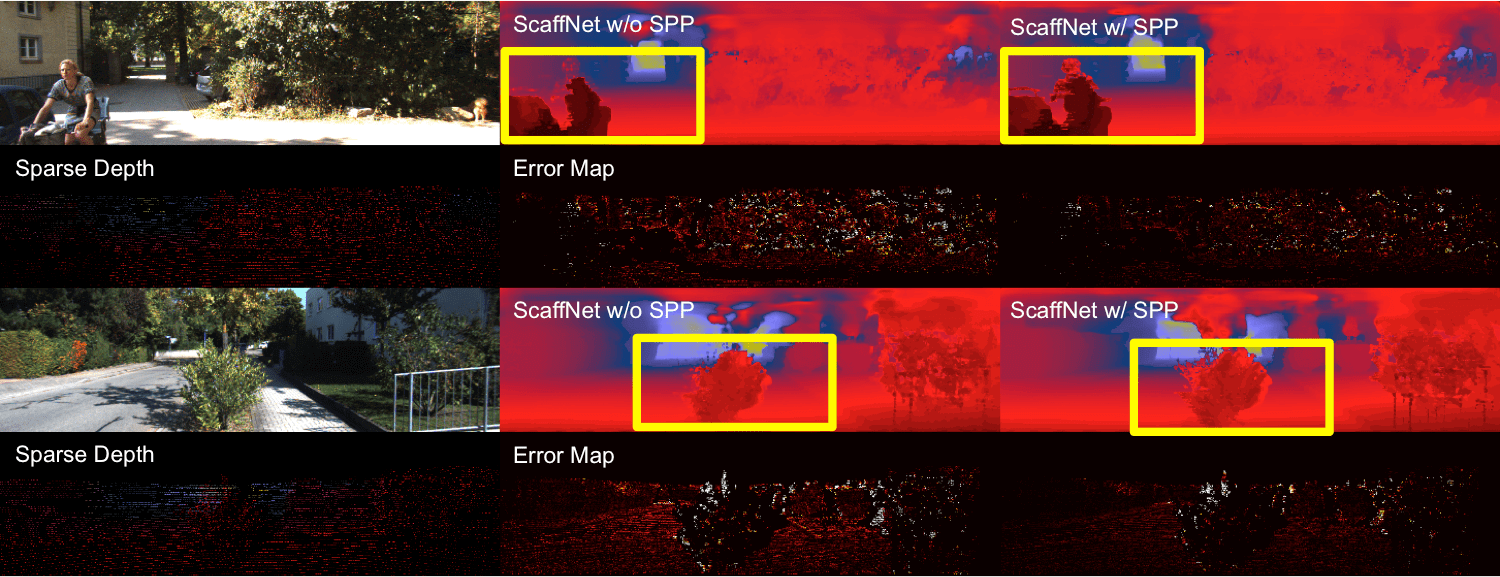}
    \caption{\textit{Qualitative ablation on the effect of SPP}. We show the benefits of leveraging SPP to increase receptive field and densify the sparse input. ScaffNet with SPP consistently outperforms the model without SPP in most regions of the error map. Using SPP, ScaffNet captures more details about the scene. For example, in the top panel, ScaffNet with SPP recovers the person while the model without SPP misses a portion of the person. In the bottom panel, we see that SPP helps retain more details about complex objects (e.g. plants and rails). Regions are highlighted in yellow for comparison.}
    \vspace{-1.5em}
    \label{fig:scaffnet_ablation}
\end{figure*}

\begin{table}[t]
    \caption{Ablation study on the KITTI validation set}
    \centering
    \setlength\tabcolsep{7.6pt}
    \begin{tabular}{l c c c c}
        \midrule 
        Method & MAE & RMSE & iMAE & iRMSE \\ 
        \midrule 
        ScaffNet w/o SPP
        & 409.93 & 1776.42 & 1.72 & 6.40 \\ 
        \midrule 
        ScaffNet w/ SPP 
        & 318.41 & 1425.53 & 1.39 & 5.01 \\ 
        \midrule 
        FusionNet ($\beta$ only) 
        & 301.08 & 1297.02 & 1.30 & 4.43 \\ 
        \midrule 
        FusionNet ($\alpha$ only) 
        & 299.90 & 1304.77 & 1.32 & 4.34 \\ 
        \midrule
        FusionNet (direct map) 
        & 292.74 & 1213.68 & 1.20 & 3.59 \\ 
        \midrule
        FusionNet (no $l_{tp}$) 
        & 297.39 & 1204.88 & 1.19 & 3.58 \\ 
        \midrule
        FusionNet
        & \textbf{286.35} & \textbf{1182.81} & \textbf{1.18} & \textbf{3.55} \\ 
        \midrule
    \end{tabular}
    \begin{tablenotes}
        Rows 1 and 2: we show a comparison between our SPP module (w/ SPP) and conventional convolutions (w/o SPP) for processing the sparse inputs. SPP improves performance across all metrics by large margins. Rows 3 to 5 and 7: we justify learning $\alpha$ and $\beta$ for refining the initial estimate $\hat{d}_{0}$. $\alpha$ and $\beta$ alone are unable to capture the scene and performs worse than direct mapping. When combined together $\alpha$ and $\beta$ surpasses direct mapping on all metrics. Rows 6 and 7: the topology prior (\eqnref{eqn:topology_prior_loss}) allows us to leverage what we have learned from synthetic data in regions that are compatible with the image, providing a consistent performance boost. Note that Our ScaffNet is using sparse-depth only.
    \end{tablenotes}
    \label{tab:kitti_validation_set_ablation}
    \vspace{-1.5em}
\end{table}

\section{Experiments and Results}
\label{sec:experiments_results}
\subsection{KITTI Depth Completion Benchmark}
We evaluate our method on the \textit{unsupervised} KITTI depth completion benchmark (outdoor driving scenarios), using metrics defined in \tabref{tab:error_metrics}. In \tabref{tab:kitti_validation_set_results}, we compare ScaffNet, which uses \textit{only} sparse depth and trained in synthetic data, to recent unsupervised methods \cite{ma2019self,wong2020unsupervised,yang2019dense} that learn dense depth from \textit{both} sparse depth and RGB images of the target (real) domain. We note that ScaffNet has never been trained on real data, yet, outperforms \cite{ma2019self,yang2019dense} by as much as 11.29\% and 8.28\%, respectively, in the MAE metric. Our findings verify that it is indeed possible to produce a decent topology estimate from only sparse inputs. More importantly, this shows that despite being trained on synthetic data, ScaffNet can bypass the domain gap and generalize to real scenarios. We note that the topology produced by ScaffNet is only an initial approximation. By fusing image information with the approximation, our FusionNet outperforms the top unsupervised depth completion methods on every metric.

Our approach also outperforms the state of the art, VGG11 \cite{wong2020unsupervised}, across every metric while having a 19.6\% parameter reduction on the KITTI depth completion benchmark testing set (\tabref{tab:kitti_testing_set_results}). We attribute part of our success to ScaffNet, whose output serves as both an initialization as well as a prior for FusionNet. On its own, ScaffNet outperforms scaffolding \cite{wong2020unsupervised} by as much as 28.22\% on the MAE metric (see \tabref{tab:kitti_validation_set_results}). We note key comparisons: in contrast to \cite{wong2020unsupervised}, we learn $\alpha$ (multiplicative scale) and $\beta$ (additive residual) around the initial approximation and hence alleviates FusionNet from having to re-learn depth, allowing us to reduce parameters while achieving better performance. Also, unlike \cite{yang2019dense}, our prior does not require an image, which is subject to domain gap between real and synthetic data. 

\begin{figure*}[th]
    \vspace{-1.5em}
    \centering
    \includegraphics[width=0.84\linewidth,height=0.488\linewidth]{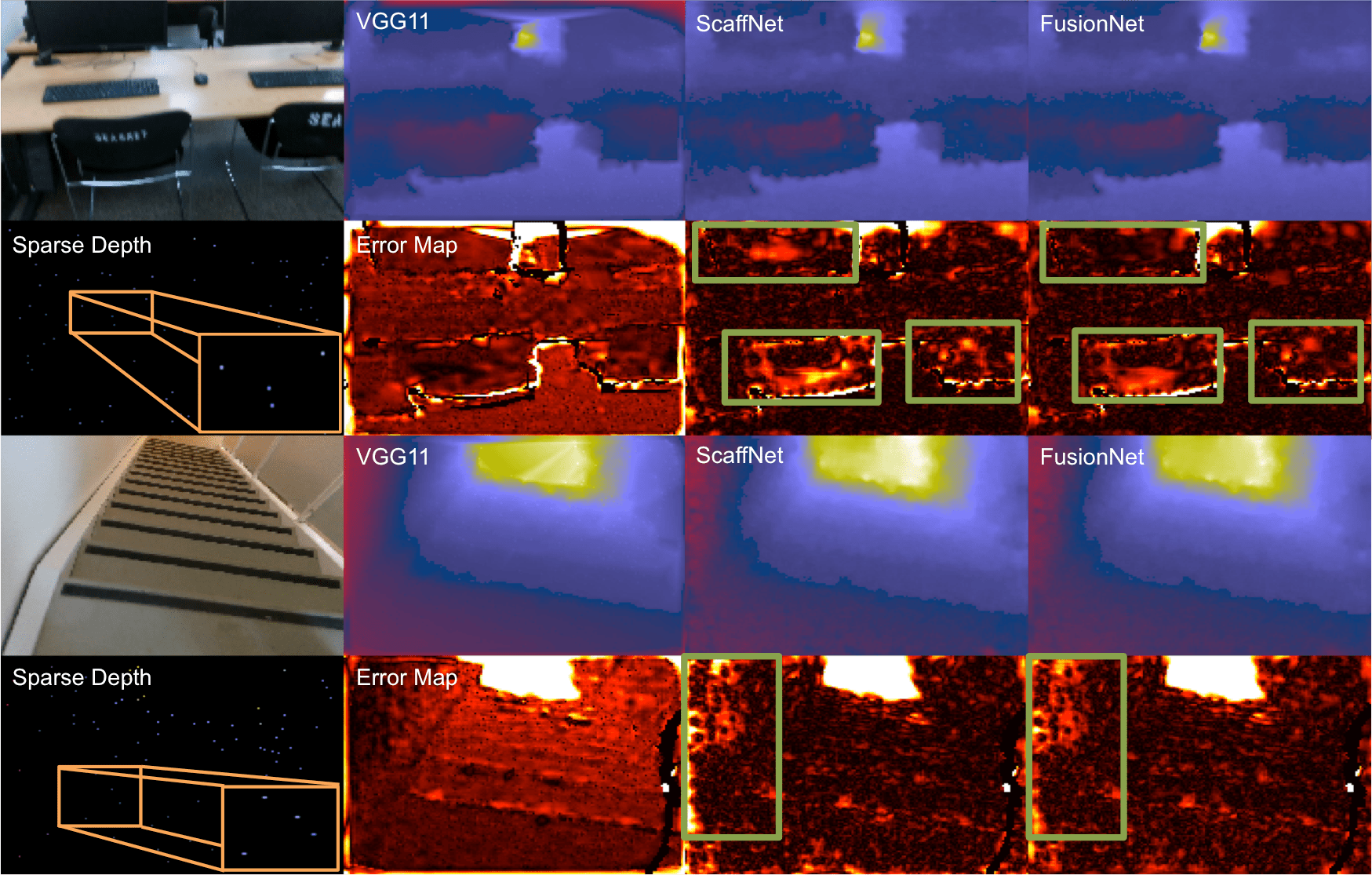}
    \caption{\textit{VOID depth completion testing set.} Although ScaffNet only uses sparse points (magnified) and is trained on synthetic data, it can still capture the topology of the scene. FusionNet approximated topology and refines the incorrect predictions (regions highlighted in green) with image information. Note:  in the bottom two rows, the initial estimate is mostly correct; hence FusionNet does not modify those regions and instead leverages it as a prior (\eqnref{eqn:topology_prior_loss}).} 
    \vspace{-1.5em}
    \label{fig:void_testing_results}
\end{figure*}

\begin{table}[th]
    \caption{Quantitative results on the VOID benchmark}
    \setlength\tabcolsep{8.3pt}
    \begin{tabular}{l c c c c}
            \midrule 
            Method & MAE & RMSE & iMAE & iRMSE \\ \midrule
            Ma \cite{ma2019self} 
            & 178.85 & 243.84 & 80.12 & 107.69 \\ \midrule 
            Yang \cite{yang2019dense} 
            & 151.86 & 222.36 & 74.59 & 112.36 \\ \midrule 
            VGG8 \cite{wong2020unsupervised} 
            & 98.45 & 169.17 & 57.22 & 115.33 \\ \midrule
            VGG11 \cite{wong2020unsupervised} 
            & 85.05 & 169.79 & 48.92 & 104.02 \\ \midrule
            Our ScaffNet
            & 70.16 & 156.99 & 42.78 & 91.48 \\ \midrule
            VGG11 + SLAM \cite{wong2020unsupervised} 
            & 73.14 & 146.40 & 42.55 & 93.16 \\ \midrule
            Our FusionNet 
            & \textbf{59.53} & \textbf{119.14} & \textbf{35.72} & \textbf{68.36} \\ \midrule
        \end{tabular}
    \begin{tablenotes}
        Results of \cite{ma2019self,wong2020unsupervised,yang2019dense} are taken from \cite{wong2020unsupervised}. Because there are many textureless regions in indoor scenes, locally, the image does not inform the scene structure. Hence, a prior informed by data is even more important. Our method outperforms all competing methods on the VOID depth completion benchmark to achieve the state of the art.
    \end{tablenotes}
    \label{tab:void_testing_set_results}
    \vspace{-1.8em}
\end{table}

To evaluate the contribution of SPP, we provide ablation study in \tabref{tab:kitti_validation_set_ablation}. As seen in row 1 and 2, by augmenting our topology estimator with SPP, we gain a 22.32\% error reduction on the MAE metric. This shows that the increase in receptive field and weighted multi-scale densification of the sparse depth inputs are important elements in tackling the sparse depth completion problem. We illustrate the benefits of our SPP module in \figref{fig:scaffnet_ablation}. The model with SPP consistently outperforms the one without and is able to retain more details about the scene with significantly lower errors.

To understand the architectural choice, we evaluated FusionNet using different output in rows 3 to 5 and 7 of \tabref{tab:kitti_validation_set_ablation}. When learning $\alpha$ (scale) and $\beta$ (residual) individually, FusionNet performs worse than directly mapping (learning from scratch) the initial approximation $\hat{d}_{0}$ and the image to dense depth. However, when combined together, $\alpha$ and $\beta$ achieves the state of the art. We note the behavior of direct mapping during early stages of training is learning to copy depth. This is because the network weights are initialized with Gaussian noises and thus the corresponding pixel in the input prediction has the highest weight on the output prediction \cite{luo2016understanding}. By learning $\alpha$ and $\beta$ around $\hat{d}_{0}$, we relieve the network of this onerous task. In regards to why $\alpha$ and $\beta$ alone cannot achieve the same performance, we believe it is related to their distribution. We observed that, when used separately, both $\alpha$ and $\beta$ have long tail distributions (to refine regions with no sparse depth) and large local disparities (e.g. object boundaries); whereas when used together, the range of their values are much smaller. We hypothesize that the sharp local changes make it harder to learn just $\alpha$ or $\beta$. 

Finally, we assess the effect of our topology prior in rows 6 and 7. Row 6 (no $l_{tp}$) omits the topology prior from the objective function (\eqnref{eqn:objective_function}). We see an overall improvement by leveraging what we have learned from synthetic datasets. This also shows that the topology obtained from synthetic data can generalize to real datasets. The proposed method (FusionNet) surpasses all  baselines and variants to justify each architectural and loss component choice.

\vspace{-0.2em}
\subsection{VOID Depth Completion Benchmark}
\label{sec:void_depth_completion_benchmark}
We provide quantitative (\tabref{tab:void_testing_set_results}) and qualitative (\figref{fig:void_testing_results}) evaluations of our approach on the indoor and outdoor VOID \cite{wong2020unsupervised} depth completion benchmark. Indoor scenes are composed of many textureless surfaces. Locally, they do not give any information about the scene structure; hence, learning a topology prior that is informed by data is even more important. The challenge here is that indoor scenes contain objects with more shape variations (from simple flat surfaces like walls to complex ones like chairs). Moreover, the density of the sparse points (tracked by VIO and SLAM systems) is $\approx$0.5\% in contrast to outdoor driving scenarios (lidar), where density is $\approx$5\% concentrated on the lower half of the image space. To demonstrate that ScaffNet can generalize even when the scene structures in the synthetic dataset is different from that of the real dataset, we trained our ScaffNet on SceneNet \cite{mccormac2016scenenet} and evaluated on VOID -- SceneNet consists of randomly arranged synthetic indoor household rooms while VOID contains real indoor and outdoor scenes of laboratories, classrooms and gardens. Despite having far fewer points, more complex geometry, and being trained on a synthetic dataset with different scene distribution, ScaffNet is still able to predict reasonable topology -- in fact, it beats all of the competing methods that are trained on real data using both image and sparse depth. This is because ScaffNet does not learn the scenes themselves, but the shapes of natural objects populating them, which allows the network to exploit the abundance of synthetic data to learn patterns from sparse points to infer dense topology. Our FusionNet further incorporates the image information into the topology estimate to achieve the state of the art, beating VGG11 \cite{wong2020unsupervised} by as much as $\approx$30\% on MAE. We also outperform their hybrid model VGG11+SLAM (which uses accurate SLAM pose instead of learning it from scratch) by $\approx$18.6\% on MAE. To show the effect of different densities of sparse inputs, we provide an ablation study in the Sec. II of Supp. Mat. -- we show that while performance does degrade with fewer points, it degrades more gracefully than \cite{wong2020unsupervised}. 

\section{Discussion}
To revisit the question, ``is it \textit{possible} to learn dense topology from sparse points?'', we have demonstrated that it is, indeed, not only possible to learn the association of sparse points with dense natural shapes, but also using \textit{only synthetic data}. While one may surmise that ScaffNet requires similar distributions of 3D scenes between synthetic and real datasets in order to generalize, we show the contrary; ScaffNet learns the shape of objects populating a scene rather than the scene itself as demonstrated in \secref{sec:void_depth_completion_benchmark}. This is especially important in the indoor settings where not only do the scene layouts vary a lot, but also consist of many textureless surfaces (for which one \textit{needs} a prior on the shapes in the scene). We must note that the topology estimated by ScaffNet is only a ``guess'' and therefore must be reconciled with the image via FusionNet (e.g. regions with very few or no sparse points, where estimates from ScaffNet are less reliable). Hence, by leveraging the topology as a prior and learning the residual over it, we allow FusionNet the freedom to amend the scene as needed. Also, we did not consider the case where the shapes in synthetic data are not representative of those in the real data. Hence, this is only the first step. We believe our findings demonstrate the benefits of leveraging synthetic data for learning topology from sparse points and motivates further exploration to incorporate the virtually unlimited amount of synthetic data into multi-sensor fusion pipelines for the 3D reconstruction task.

\bibliographystyle{ieee}
\bibliography{condensebib}

\vspace{2em}

\begin{center}
    {\LARGE{\textbf{Supplementary Materials}}}
\end{center}

\vspace{1em}

\begin{appendices}

\textbf{Summary of content:} In \secref{sec:ablation_study_spp}, we begin by validating the claim made in Sec. III-A of the main text regarding the sparsity in the feature maps produced by early convolutional layers without Spatial Pyramid Pooling. To demonstrate the effect of Spatial Pyramid Pooling, we show, in \figref{fig:spatial_pyramid_pooling_ablation}, a visualization of this phenomenon and how augmenting the network with a Spatial Pyramid Pooling module can produce much denser feature maps. Our choice of kernel sizes for Spatial Pyramid Pooling is detailed in \secref{sec:ablation_study_spp_kernel_sizes}. In \secref{sec:ablation_study_spp_performance}, we show (i) additional ablation studies on the performance benefits of Spatial Pyramid Pooling, (ii) discuss how Spatial Pyramid Pooling balances the trade-off between level of density in the extracted features and the level of detail that they capture in \secref{sec:ablation_study_spp_tradeoff}, and (iii) how our variant of Spatial Pyramid Pooling is different from previous works in \secref{sec:ablation_study_spp_differences}. In \secref{sec:density_ablation_studies}, we examine the impact of various levels of input sparse depth density for both ScaffNet and FusionNet (\tabref{tab:void_density_ablation} and \figref{fig:void_density_ablation}, \ref{fig:scaffnet_density_comparison_void}). In \secref{sec:sparse_point_sensitivity_studies}, we examine ScaffNet's sensitivity to different distributions of sparse depth sampling strategies.

In \secref{sec:procedures_training_scaffnet_fusionnet}, we shift our focus to examining the full system (ScaffNet with FusionNet) and show the different options for training them (either separately or jointly). We provide quantitative results, in \tabref{tab:ablation_training_method_kitti_validation}, to demonstrate effects of separate and joint training to justify our choice in training procedure and our use of synthetic data. To understand the effect of domain gap on performance, in \secref{sec:generalization_studies}, we evaluate ScaffNet and FusionNet on a dataset where camera setup and scene distribution differs from that of the training set.

Finally, in \secref{sec:unsupervised_kitti_benchmark}, we further show quantitative and qualitative comparisons with other unsupervised methods on the KITTI depth completion benchmark. We also compare our approach to supervised methods and show that our method is closing the gap between supervised and unsupervised learning paradigms. In \figref{fig:kitti_benchmark_screenshot}, we show screenshots of the unsupervised approaches with their respective ranks on the KITTI benchmark webpage. In \secref{sec:network_architectures}, we detail the network architectures of our topology estimator (ScaffNet) and depth-RGB image fusion network (FusionNet), both were specifically designed to be light-weight with the intention of being able to be deployed on standard embedded systems for real-time applications.

\section{Ablation Study on the effects of \\ Spatial Pyramid Pooling}
\label{sec:ablation_study_spp}
\subsection{Regarding its effect on feature maps}
\begin{figure*}[th]
    \centering
    \includegraphics[width=1.00\linewidth]{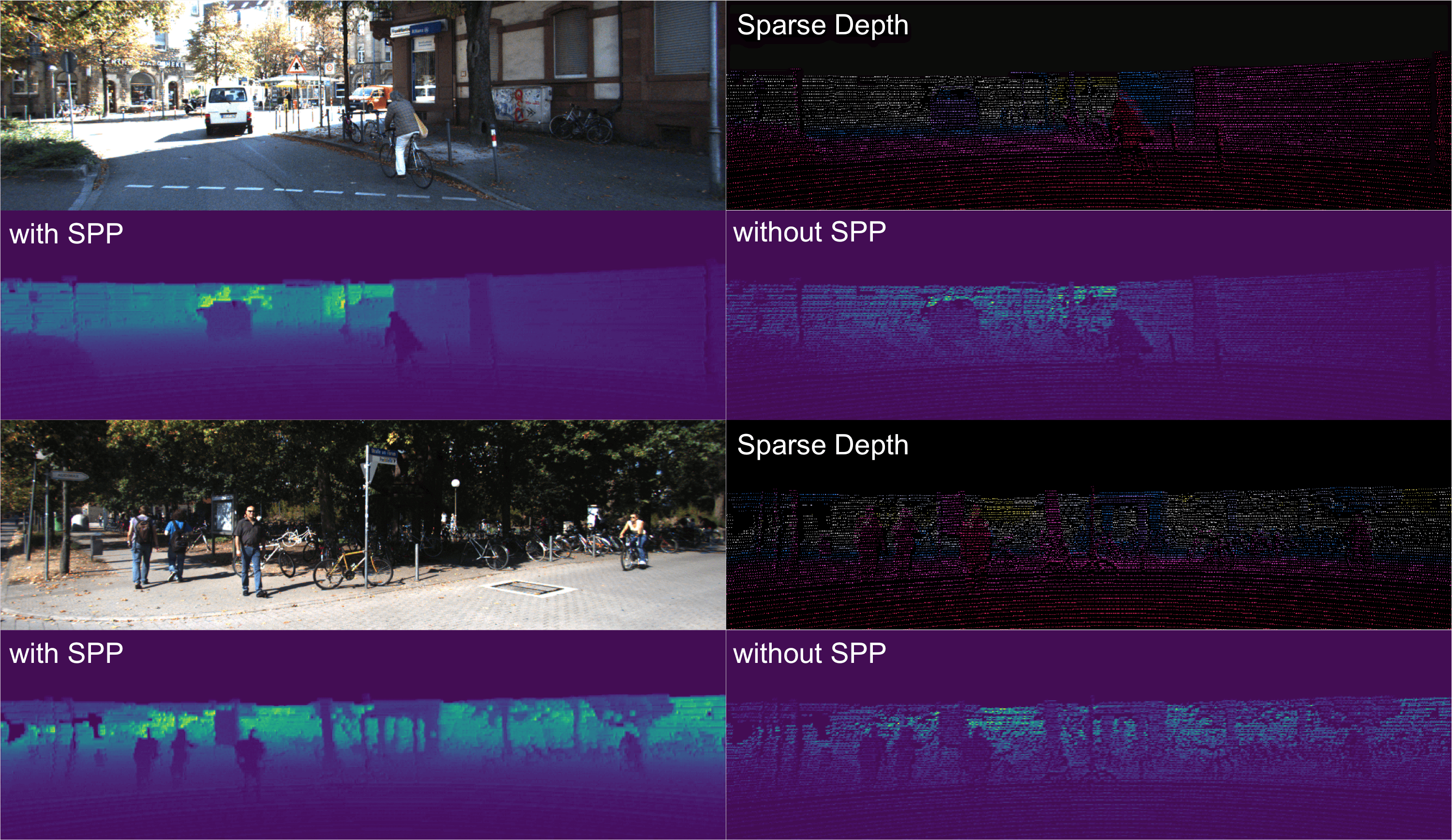}
    \caption{\textit{A comparison of feature maps produced by ScaffNet with and without Spatial Pyramid Pooling (SPP)} (best viewed $5\times$ and in color). Hidden layer outputs (feature maps) after 5 $\times$ 5 convolutional layers with and without SPP. The feature maps produced by ScaffNet without SPP are sparse and resembles the input sparse depth, which illustrates our claim in the main paper -- neurons are not activated because of the missing sparse depth measurements. In contrast, when using SPP, ScaffNet produces a much denser representation.}
\label{fig:spatial_pyramid_pooling_ablation}
\end{figure*}

The challenge of sparse depth completion is precisely the sparsity. Because of the numerous ``holes'' (or zeroes) in the sparse input, the activations of the earlier convolutions layers tend to be zero as well. Therefore, much of the earlier layers are dedicated to ``densifying'' the feature maps. This is illustrated in the right column of \figref{fig:spatial_pyramid_pooling_ablation}. Our goal is to enable better learning by generating a denser representation. To this end, we proposed augmenting the topology estimator network (ScaffNet, Sec. III-A of main text) with our Spatial Pyramid Pooling module, where multiple max pools of different scales are performed on the sparse input for an increase in receptive field and input densification. In effect, the features generated from Spatial Pyramid Pooling become much denser than those from simple convolutions. This, in turn, allows the activations of the subsequent convolutional layers in the encoder to be dense as well. We illustrate this phenomenon in \figref{fig:spatial_pyramid_pooling_ablation} where we show that the feature maps of ScaffNet \textit{without} the use of Spatial Pyramid Pooling is still sparse and resembles the input sparse depth; whereas, the feature maps of ScaffNet \textit{with} Spatial Pyramid Pooling is much denser in comparison. By providing the encoder with dense outputs from Spatial Pyramid Pooling, we, not only, relieve the encoder from the onerous task of propagating the sparse signal spatially, but also provide larger receptive field and context to the subsequent layers. 

\begin{table}[t]
    \caption{Comparison of ScaffNet Pooling Kernel Sizes on VOID}
    \setlength\tabcolsep{7.5pt}
    \centering
    \begin{tabular}{l c c c c c}
            \midrule 
            Pool Sizes & MAE & RMSE & iMAE & iRMSE \\ 
            \midrule
            none & 100.75 & 242.27 & 71.32 & 191.60 \\ 
            \midrule
            5 & 102.82 & 217.38 & 69.05 & 195.51 \\ 
            \midrule
            5, 7 & 86.44 & 173.06 & 56.68 & 174.86  \\
            \midrule
            5, 7, 9 & 81.48 & 169.24 & 51.76 & 140.33  \\
            \midrule
            5, 7, 9, 11 & 77.104 & \textbf{154.604} & 46.886 & 121.256 \\
            \midrule
            \textbf{5, 7, 9, 11, 13} & \textbf{70.16} & 156.99 & \textbf{42.78} & \textbf{91.48} \\
            \midrule
            3, 5, 7, 9, 11, 13 & 71.49 & 154.91 & 44.01 & 91.65 \\
            \midrule
            5, 7, 9, 11, 13, 15 & 72.55 & 159.31 & 44.67 & 99.15 \\
            \midrule
            5, 7, 9, 11, 13, 15, 17 & 71.93 & 156.09 & 46.15 & 97.50 \\
            \midrule
    \end{tabular}
    \begin{tablenotes}
        ScaffNet is trained, using various max pooling kernel sizes, on SceneNet and evaluated on VOID. When using just a 5 by 5 kernel (row 2), performance is similar to not using SPP at all (row 1, none). Including 7 by 7 max pooling increases performance across all metrics. We observe consistent performance gain as we increase kernel size up to 13 by 13 (row 6, the proposed method). Adding more max pooling layers with larger kernel size (e.g. 15 in row 8, 17 in row 9) does not increase performance. Adding small kernel size e.g. (3, in row 7) also does not increase performance.
    \end{tablenotes}
\label{tab:scaffnet_pooling_kernel_sizes_void}
\vspace{-2em}
\end{table}

\subsection{Regarding the choice of kernel sizes}
\label{sec:ablation_study_spp_kernel_sizes}
For Spatial Pyramid Pooling, we choose the following kernel sizes for max pooling: $5 \times 5$, $7 \times 7$, $9 \times 9$, and $11 \times 11$ for KITTI, and $5 \times 5$, $7 \times 7$, $9 \times 9$, $11 \times 11$, and $13 \times 13$ for VOID. These settings are chosen empirically and we validate our choice of kernel sizes used in ScaffNet in \tabref{tab:scaffnet_pooling_kernel_sizes_void}, where we show quantitative results on various kernel sizes used in Spatial Pyramid Pooling. With small pool sizes e.g. $5 \times 5$ (row 2), there is only small performance gain over not using SPP (row 1, none). This is because the feature maps will still be sparse and therefore the network must dedicate its early layers to propagating the signal. As soon as we add a $7 \times 7$ max pool (row 2), we observe a performance boost across all metrics; this is consistently the case as we add more max pool layers with larger kernel sizes. The best performing combination is the proposed setting, (5, 7, 9, 11, 13, row 6). Increasing the number max pooling layers with larger kernel sizes ($15 \times 15$, row 8 and $17 \times 17$, row 9) does not increase performance; neither does adding max pooling layer with smaller kernel size ($3 \times 3$, row 7). This is likely because local information captured by a $3 \times 3$ can likely also be captured by the combination of $5 \times 5$ max pooling and the original input.

\subsection{Regarding its effect on performance} 
\label{sec:ablation_study_spp_performance}
To understand the benefits of using Spatial Pyramid Pooling, we show an ablation study on its impact on ScaffNet (when augmented with this module) and also its effect on later stages of inference (FusionNet) in both outdoors (KITTI, \tabref{tab:kitti_spatial_pyramid_pooling_ablation}) and indoors (VOID, \tabref{tab:void_spatial_pyramid_pooling_ablation}) scenarios. 
Overall, the performance benefits are apparent in \tabref{tab:kitti_spatial_pyramid_pooling_ablation} and \ref{tab:void_spatial_pyramid_pooling_ablation}, where both ScaffNet  augmented with Spatial Pyramid Pooling and its associated FusionNet (marked with w/ SPP) outperform their variants without the module (marked with w/o SPP) by large margins across all metrics. Because ScaffNet consistently performs worse without Spatial Pyramid Pooling, FusionNet is given a lower quality initial topology; therefore, FusionNet like-wise performs worse -- justifying the use of Spatial Pyramid Pooling in ScaffNet. This ablation study also shows that errors are propagated downstream. However, while errors do get introduced to FusionNet when using a ScaffNet without Spatial Pyramid Pooling, this is not a point of failure as FusionNet (w/o SPP) is still able to amend the mistakes and improve the reconstruction. We also note that our ScaffNet without Spatial Pyramid Pooling, in fact, still outperforms \cite{ma2019self,yang2019dense} in \tabref{tab:void_spatial_pyramid_pooling_ablation}. We note it is possible to use ScaffNet to obtain dense topology from sparse depth to benefit existing supervised and unsupervised depth completion methods including, but not limited to \cite{chen2019learning,ma2019self,shivakumar2019dfusenet,qiu2019deeplidar,uhrig2017sparsity,wong2020unsupervised,xu2019depth,yang2019dense,zhang2018deep}.

\begin{table}[t]
    \centering
    \setlength\tabcolsep{8pt}
    \caption{Ablation study on the effect of Spatial Pyramid Pooling}
    \begin{tabular}{l c c c c}
        \midrule 
        Method & MAE & RMSE & iMAE & iRMSE \\ \midrule
        Scaffolding \cite{wong2020unsupervised}   
        & 443.57 & 1990.68 & 1.72 & 6.43 \\ \midrule
        Our ScaffNet w/o SPP
        & 409.93 & 1776.42 & 1.72 & 6.40 \\ \midrule 
        Our ScaffNet w/ SPP
        & 318.41 & 1425.53 & 1.39 & 5.01 \\ \midrule 
        Ma \cite{ma2019self} 
        & 358.92 & 1384.85 & 1.60 & 4.32 \\ \midrule
        Yang \cite{yang2019dense}
        & 347.17 & 1310.03  & n/a & n/a \\ \midrule
        VGG8 \cite{wong2020unsupervised}   
        & 308.81 & 1230.85 & 1.29 & 3.84 \\ \midrule
        VGG11 \cite{wong2020unsupervised} 
        & 305.06 & 1239.06 & 1.21 & 3.71 \\ \midrule
        Our FusionNet w/o SPP
        & 306.54 & 1219.92 & 1.24 & 3.65 \\ \midrule
        Our FusionNet w/ SPP
        & \textbf{286.35} & \textbf{1182.81} & \textbf{1.18} & \textbf{3.55} \\ \midrule
    \end{tabular}
    \begin{tablenotes}
        \textit{Ablation study on the effect of Spatial Pyramid Pooling on KITTI validation set}. Results of \cite{ma2019self,yang2019dense,wong2020unsupervised} are directly taken from their papers. Our ScaffNet w/o SPP omits the Spatial Pyramid Pooling (SPP) module. Our FusionNet w/o SPP uses ScaffNet w/o SPP as the initial topology estimator. Results of our ScaffNet with SPP consistently impoves its variant without the module. Similarly, because ScaffNet with SPP provides FusionNet with more accurately estimated topology, FusionNet like-wise improves.
    \end{tablenotes}
\label{tab:kitti_spatial_pyramid_pooling_ablation}
\end{table}

\subsection{Regarding the trade-off between detail and density}
\label{sec:ablation_study_spp_tradeoff}
We note that for the purpose of densification, one may just use a single max pool layer with a large kernel size. However, such a max pool layer would decimate the details of the sparse input. One may also choose to leverage heuristics such as nearest neighbor and local smoothness \cite{wong2020unsupervised} to interpolate depth between sparse points. However, when the nearest neighboring points are far away (both in 3D world or 2D image space), the plane interpolated between the points may contain incorrect values as the points may violate the local connectivity assumption. Hence, we use multiple kernel sizes for our max pool layers to capture local fine details (with small kernel sizes) and global denser structures (with large kernel sizes). To determine the trade-off between details and density, we leverage synthetic data to train three $1 \times 1$ convolutional layers to weight the output of the max pool layers. For network structure details, please see \secref{sec:network_architectures}.

\begin{figure*}[ht]
    \centering
    \includegraphics[width=0.90\linewidth]{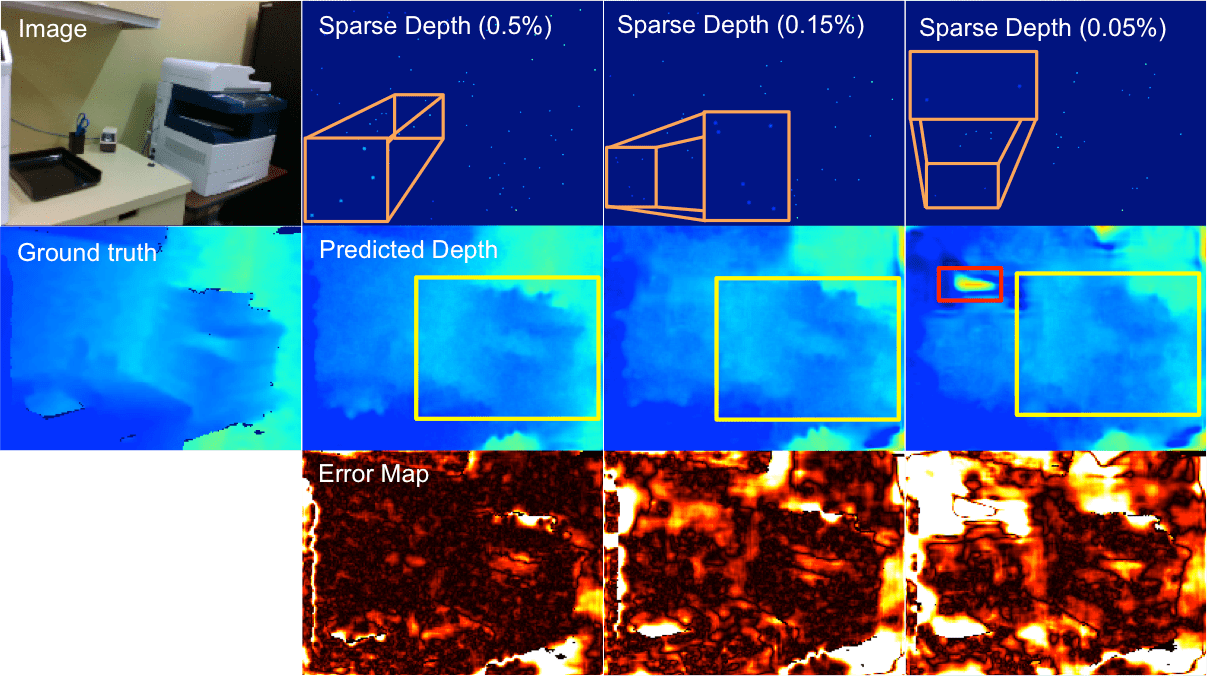}
    \caption{\textit{Qualitative comparison of ScaffNet predictions across 0.5\%, 0.15\%, and 0.05\% densities on VOID.} Orange bounding boxes show the magnified sparse depth. Because ScaffNet predictions are only informed by sparse points, performance degrades as density decreases. The shape of the printer (highlighted in yellow), loses its structure when density decreases from 0.5\% to 0.15\% and 0.05\%. At 0.05\%, ScaffNet starts to produce some artifacts (``holes'', highlighted in red) because there are large missing regions.}
\label{fig:scaffnet_density_comparison_void}
\vspace{-1em}
\end{figure*}

\begin{table}[t]
    \setlength\tabcolsep{8pt}
    \centering
    \caption{Ablation study on the effect of Spatial Pyramid Pooling}
    \begin{tabular}{l c c c c}
        \midrule 
        Method & MAE & RMSE & iMAE & iRMSE \\ \midrule
        Ma \cite{ma2019self} 
        & 198.76 & 260.67 & 88.07 & 114.96 \\ \midrule 
        Yang \cite{yang2019dense} 
        & 151.86 & 222.36 & 74.59 & 112.36 \\ \midrule 
        Our ScaffNet w/o SPP
        & 100.75 & 242.27 & 71.32 & 191.60 \\ \midrule
        VGG8 \cite{wong2020unsupervised} 
        & 98.45 & 169.17 & 57.22 & 115.33 \\ \midrule
        VGG11 \cite{wong2020unsupervised} 
        & 85.05 & 169.79 & 48.92 & 104.02 \\ \midrule
        Our FusionNet w/o SPP
        & 77.62 & 140.36 & 51.58 & 91.84 \\ \midrule
        Our ScaffNet w/ SPP
        & 70.16 & 156.99 & 42.78 & 91.48 \\ \midrule
        VGG11 + SLAM \cite{wong2020unsupervised} 
        & 73.14 & 146.40 & 42.55 & 93.16 \\ \midrule
        Our FusionNet w/ SPP
        & \textbf{59.53} & \textbf{119.14} & \textbf{35.72} & \textbf{68.36} \\ \midrule
    \end{tabular}
    \begin{tablenotes}
        \textit{Ablation study on the effect of Spatial Pyramid Pooling on VOID depth completion benchmark using $\approx$1500 points ($\approx$0.5\% density)}. Results of \cite{ma2019self,yang2019dense,wong2020unsupervised} are taken from \cite{wong2020unsupervised}. Results of w/o SPP consistently performs worse than our model with SPP (marked with w/ SPP). We note that while errors do propagate from ScaffNet to FusionNet, FusionNet is able to amend them as see in the entry ``Our FusionNet w/o SPP''. We also note that our ScaffNet w/o SPP still outperforms \cite{ma2019self,yang2019dense}.
    \end{tablenotes}
\label{tab:void_spatial_pyramid_pooling_ablation}
\vspace{-1em}
\end{table}

\subsection{Differences from previous use cases}
\label{sec:ablation_study_spp_differences}
While variants of Spatial Pyramid Pooling have been employed in other problems such as classification and stereo matching, our use of Spatial Pyramid Pooling is unique. \cite{he2015spatial} introduced Spatial Pyramid Pooling to ensure the same size feature maps are maintained when different size of inputs are fed through the network. Unlike us, \cite{he2015spatial} does not re-weight the features and directly feed max pooled results to fully connected layers. \cite{chang2018pyramid} used Spatial Pyramid (Average) Pooling with large kernels to create ``region-level'' features for increasing receptive field. In contrast to \cite{chang2018pyramid}, we use max pooling to avoid loss of local detail from averaging large regions. Also unlike our use case in the depth completion problem, in classification and stereo, the inputs are \textit{dense}; whereas, ours are \textit{sparse}. Hence, we leverage the assumption that surfaces exhibit local smoothness and connectivity and perform max pooling with large kernels over local regions to get coarse local representations and max pooling with small kernels to retain detail. To balance the trade-off between detail and density, we re-weight the pooled features with $1 \times 1$ convolutions. This design not only allows us to obtain a denser representation, but also to obtain larger receptive field (with similar motivation as \cite{chang2018pyramid}) -- differentiating our variant of Spatial Pyramid Pooling from previous works.

\begin{figure}[t]
    \centering
    \includegraphics[width=1.00\linewidth]{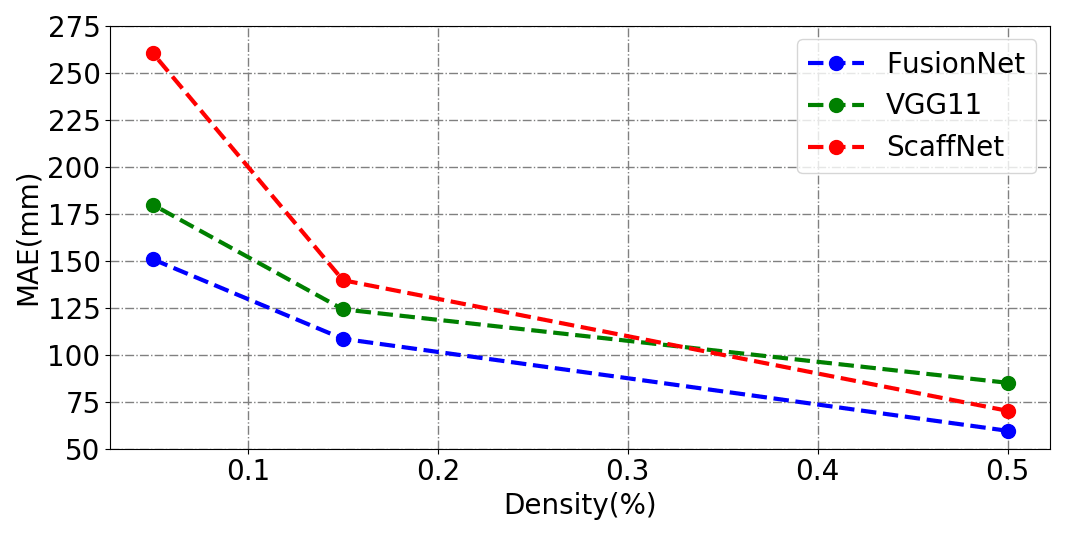}
    \caption{\textit{Plot of MAE at different densities on the VOID depth completion benchmark.} Results of \cite{wong2020unsupervised} are taken directly from their paper. The three density levels examined are $\approx$0.5\%, $\approx$0.15\%, and $\approx$0.05\%, which corresponds to $\approx$1500, $\approx$500, and $\approx$150 points, respectively. We show the trends of the MAE metric for ScaffNet, FusionNet and VGG11 \cite{wong2020unsupervised}. While ScaffNet beats VGG11 at the highest density ($\approx$0.5\%), with fewer points, ScaffNet performance degrades more quickly than FusionNet and VGG11 \cite{wong2020unsupervised}. This is because ScaffNet only takes sparse points as input (i.e without RGB image) and therefore cannot reliably infer the scene if there are very few or no sparse points. The improvement from ScaffNet to FusionNet is the effect of the errors amended by FusionNet.}
\label{fig:void_density_ablation}
\vspace{-0.5em}
\end{figure}

\begin{table}[t]
    \caption{Ablation study on various densities of sparse inputs}
    \centering
    \setlength\tabcolsep{11pt}
    \begin{tabular}{l c c c c}
        \midrule 
        & \multicolumn{4}{c}{$\approx$0.15\% density} \\
        \midrule 
        Method & MAE & RMSE & iMAE & iRMSE \\
        \midrule
        Our ScaffNet 
        & 139.60 & 308.47 & 71.50 & 151.49 \\ 
        \midrule 
        VGG11 \cite{wong2020unsupervised} 
        & 124.11 & 217.43 & 66.95 & 121.23 \\ 
        \midrule
        Our FusionNet 
        & \textbf{108.44} & \textbf{195.82} & \textbf{57.52} & \textbf{103.33} \\ 
        \midrule
        \midrule
        & \multicolumn{4}{c}{$\approx$0.05\% density} \\
        \midrule
        Method & MAE & RMSE & iMAE & iRMSE \\ 
        \midrule
        Our ScaffNet 
        & 260.13 & 531.69 & 115.21 & 218.27 \\ 
        \midrule
        VGG11 \cite{wong2020unsupervised}
        & 179.66 & 281.09 & 95.27 & 151.66 \\ 
        \midrule
        Our FusionNet 
        & \textbf{150.65} & \textbf{255.08} & \textbf{80.79} & \textbf{133.33} \\ 
        \midrule
    \end{tabular}
    \begin{tablenotes}
    Results of \cite{wong2020unsupervised} are taken from their papers. We examined two other density levels, $\approx$0.15\%, and $\approx$0.05\%, in addition to $\approx$0.5\% density shown in the official benchmark (see \tabref{tab:void_spatial_pyramid_pooling_ablation}), which corresponds to $\approx$500, and $\approx$150 points, respectively. The performance of ScaffNet and FusionNet  decreased (as expected) proportional to the density, but ScaffNet decreases at a faster rate with fewer points, especially at 0.05\% density. This is because ScaffNet needs to infer topology from only $\approx$150 points without the help of an image. However, with an input of $\approx$0.15\% density, our approach still produces reasonable results, with numbers similar to that of \cite{ma2019self,yang2019dense,wong2020unsupervised} using $\approx$0.5\% input density (see \tabref{tab:void_spatial_pyramid_pooling_ablation}).
    \end{tablenotes}
\label{tab:void_density_ablation}
\vspace{-1em}
\end{table}

\section{Ablation Studies on Sparse Inputs with Various Density Levels}
\label{sec:density_ablation_studies}
To understand the effect of sparse inputs of various density levels, we also show an quantitative study in \tabref{tab:void_density_ablation} and \figref{fig:void_density_ablation}, \ref{fig:scaffnet_density_comparison_void}). \figref{fig:void_density_ablation} compares the MAE metric between ScaffNet, FusionNet and VGG11 \cite{wong2020unsupervised} across three density levels: $\approx$0.5\%, $\approx$0.15\%, and $\approx$0.05\%, which corresponds to $\approx$1500, $\approx$500, and $\approx$150 points, respectively. Each model is trained on $\approx$0.5\% and tested on each density level. While ScaffNet beats VGG11 at the highest density ($\approx$0.5\%), with fewer points, ScaffNet performance degrades at a much faster rate than FusionNet and VGG11 \cite{wong2020unsupervised}. This is because ScaffNet only takes sparse points as input (i.e without RGB image) and therefore cannot reliably estimate the topology when there are very few or no sparse points. Even though ScaffNet has comparable performance as many previous methods that use both sparse depth and image as input, this is precisely why FusionNet is critical in the success of our method -- by performing cross-modal validation using the image and predicted topology to amend the incorrect predictions.

\tabref{tab:void_density_ablation} shows that FusionNet is the best performing method across all metrics at every density level. As expected, performance of both ScaffNet and FusionNet degrade with lower density; however, we note that at $\approx$0.15\% density, both ScaffNet and FusionNet are still comparable with the performance of methods at $\approx$0.5\% density (see \tabref{tab:void_spatial_pyramid_pooling_ablation}). This shows the effectiveness of our method even at low density levels, which is a common scenario for indoor scenes (e.g. features of SLAM/VIO systems where points tracked must be visually discriminative and hence generally comprise of a small set). While our method does degrade with density (as do all known depth completion methods), we degrade more gracefully than \cite{wong2020unsupervised}. We also note that the amount of ScaffNet performance degradation with respect to density is similar for all metrics. % We also note that the amount performance of ScaffNet degradation with respect to density is similar for all metrics. 

\section{Sensitivity Studies on ScaffNet for \\ Sparse Inputs with Sampling Strategies}
\label{sec:sparse_point_sensitivity_studies}
There exists different sparse point sampling strategies (e.g. horizontal scanning from lidar in KITTI, corner detection in VOID, or even random uniform sampling). In this section, we study ScaffNet's sensitivity to different sparse depth distributions and whether it is necessary to match the distribution to achieve good performance.

ScaffNet is not too sensitive to the mismatch of sparse depth distribution between synthetic and real data. We demonstrate this in \tabref{tab:comparison_sampling_strategies_void} where we train ScaffNet on SceneNet with sparse depth points uniformly randomly sampled (row 1) and evaluated it on VOID where sparse points are from VIO using corner-base feature detector. Even though they are trained on different distributions, ScaffNet can still generalized to VOID. As expected, ScaffNet trained on SceneNet with points chosen based on a corner detector (row 3) can improve performance on MAE and RMSE and is comparable on iMAE and iRMSE. FusionNet trained using a frozen ScaffNet pretrained with uniform sampling (row 2) have little performance difference compared to FusionNet trained using a frozen ScaffNet pretrained with sparse points constructed using a corner detector (row 4).

\begin{table}[t]
    \caption{Comparison of sampling strategies on the VOID benchmark}
    \setlength\tabcolsep{4pt}
    \centering
    \begin{tabular}{l c c c c c c}
            \midrule 
            & \multicolumn{2}{c}{ScaffNet trained with} & \multicolumn{4}{c}{Error metrics} \\
            Method & Uniform & Corner & MAE & RMSE & iMAE & iRMSE \\
            \midrule
            ScaffNet & \checkmark & & 76.44 & 177.63 & 40.565 & 92.589 \\
            \midrule
            FusionNet & \checkmark & & 60.393 & 126.457 & \textbf{33.237} & 69.166 \\
            \midrule
            ScaffNet & & \checkmark & 70.16 & 156.99 & 42.78 & 91.48 \\
            \midrule
            FusionNet & & \checkmark & \textbf{59.53} & \textbf{119.14} & 35.72 & \textbf{68.36} \\ \midrule
    \end{tabular}
    \begin{tablenotes}
        Comparison of ScaffNet and FusionNet trained with different sampling strategies and evaluated on the VOID dataset (points tracked by VIO with corner based feature detector). \textbf{Rows 1}: ScaffNet is trained on SceneNet where points are sampled uniformly randomly. \textbf{Row 2}: FusionNet is trained on VOID using a ScaffNet from row 1 with weights frozen. \textbf{Rows 3}: ScaffNet is trained on SceneNet where points are chosen by running Harris corner detector \cite{harris1988combined} on the corresponding image. \textbf{Row 4}: FusionNet is trained on VOID using the ScaffNet from row 3 with weights frozen. Despite being trained a synthetic dataset with points sampled uniformly, ScaffNet can generalize to real dataset with points chosen based on corner detection. Both FusionNets likewise have similar performance even though their respective ScaffNet is trained on different sparse point distributions.
    \end{tablenotes}
\vspace{-1em}
\label{tab:comparison_sampling_strategies_void}
\end{table}

\begin{figure}[t]
    \centering
    \includegraphics[width=0.70\linewidth]{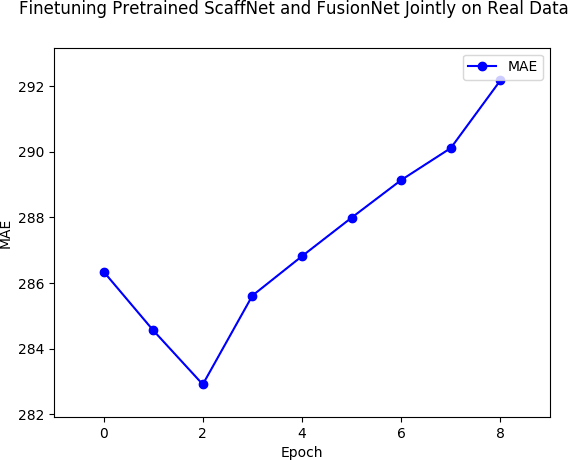}
    \caption{\textit{Finetuning pretrained ScaffNet and FusionNet on real data.} While we can get some performance gain if we were to jointly finetune ScaffNet (trained on synthetic data) and FusionNet (trained on real data with frozen ScaffNet) on real data, performance degrades after 2 epochs. This is because the image reconstruction term $l_{ph}$ is susceptible to transparent, specular surfaces. Hence, the gradients computed for these terms, backpropagated through FusionNet to ScaffNet, will destroy the topology information learned from synthetic data in ScaffNet.}
\label{fig:plot_mae_finetune_scaffnet_fusionnet}
\end{figure}

\begin{table}[t]
    \caption{Comparison of training methods on KITTI}
    \setlength\tabcolsep{5pt}
    \centering
    \begin{tabular}{c c c c c c c c}
        \midrule 
        \multicolumn{2}{c}{ScaffNet} & FusionNet & \multicolumn{4}{c}{Error metrics} \\
        \midrule
        Pretrain & Freeze & Pretrain & MAE & RMSE & iMAE & iRMSE \\ 
        \midrule
        % ScaffNet (scratch), FusionNet (scatch)	
        & & & 327.74 & 1283.02 & 1.34 & 3.79 \\
        \midrule
        % ScaffNet (pretrain), FusionNet (scatch)
        \checkmark & & & 305.90 & 1249.64 & 1.25 & 3.61 \\
        \midrule
        % ScaffNet (pretrain, frozen), FusionNet (scatch)
        \checkmark & \checkmark & & 286.35 & 1182.81 & 1.18 & 3.55 \\
        \midrule
        % ScaffNet (pretrain), FusionNet (pretrain), finetune 
        \checkmark & & \checkmark & \textbf{282.91} & \textbf{1176.09} & \textbf{1.17} & \textbf{3.54} \\
        \midrule
    \end{tabular}
    \begin{tablenotes}
        We compare different methods for training ScaffNet and FusionNet on the KITTI validation set. For ScaffNet: pretrain option denotes pretraining on synthetic data (Virtual KITTI) before training FusionNet, freeze denotes weights are frozen when training FusionNet. For FusionNet: pretrain option denotes pretraining on real data (KITTI) using a pretrained frozen ScaffNet. \textbf{Row 1}: when both ScaffNet and FusionNet are jointly trained from scratch on real data, performance is the worst because it cannot leverage \textit{side information} e.g. topology from synthetic data. \textbf{Row 2}: pretraining ScaffNet on synthetic data and jointly training ScaffNet and FusionNet on real data performs better than if ScaffNet was trained from scratch. \textbf{Row 3 (the proposed method)}: pretraining Scaffnet on synthetic data and freezing its weights while training FusionNet performs better than jointly training. \textbf{Row 4}: pretraining ScaffNet on sythetic data, freezing its weights to train FusionNet, and finetuning both jointly on real data performs slightly better than the proposed method.
    \end{tablenotes}
    \vspace{-0.5em}
    \label{tab:ablation_training_method_kitti_validation}
\end{table}

\begin{figure*}[t]
    \centering
    \includegraphics[width=1.00\linewidth]{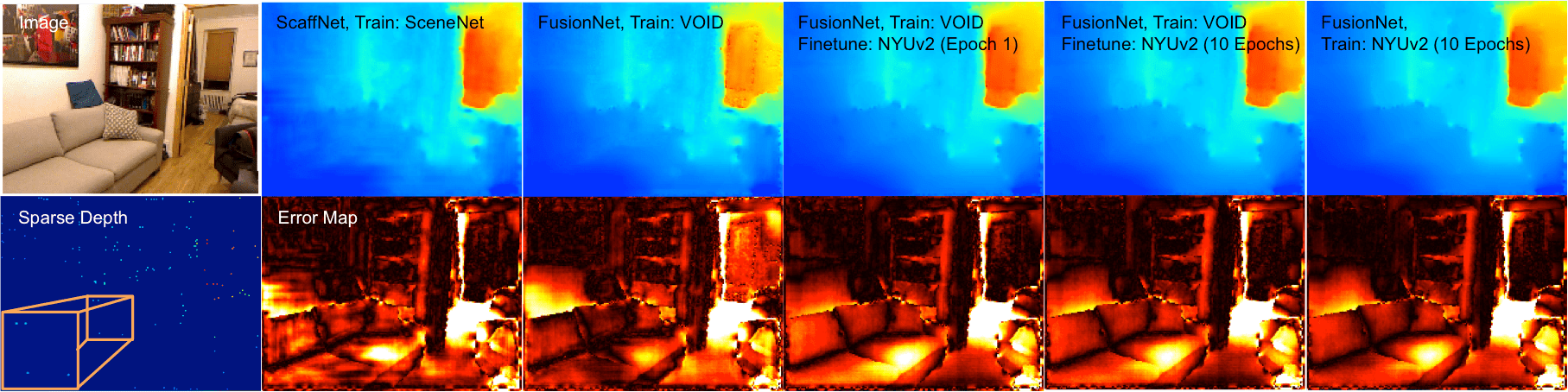} 
    \includegraphics[width=1.00\linewidth]{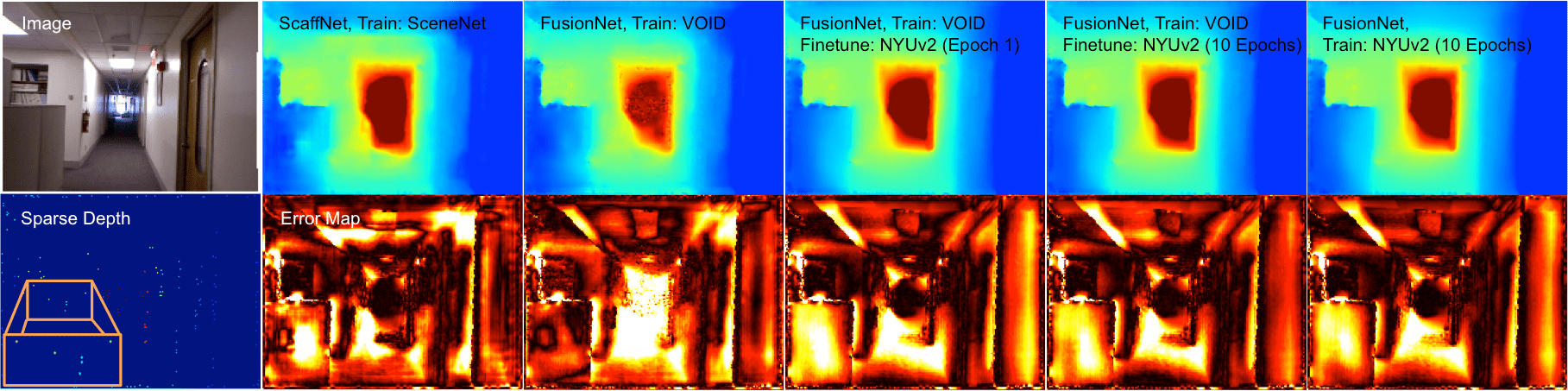}
    \caption{\textit{Qualitative evaluation on the NYUv2 test set.} Left to right, \textbf{Column 1}: Input image and sparse depth; orange bounding boxes show the magnified sparse depth. \textbf{Column 2}: Despite being trained only on synthetic data, ScaffNet is able to generalize to NYUv2. \textbf{Column 3}: FusionNet trained on VOID performs worse than ScaffNet because of the domain gap. \textbf{Column 4}: FusionNet trained on VOID recovers performance quickly by adapting to the camera and scenes of NYUv2 when finetuned for just 1 epoch. \textbf{Column 5, 6}: When finetuned longer on NYUv2, FusionNet trained on VOID performs comparably to FusionNet trained on NYUv2.}
\label{fig:nyuv2_test_set}
\end{figure*}

\section{Different Procedures for Training \\ ScaffNet and FusionNet}
\label{sec:procedures_training_scaffnet_fusionnet}
There are several ways to train the full system (ScaffNet and FusionNet) either by training them separately or jointly. Our training procedure is as follows: (i) train ScaffNet on synthetic data, (ii) freeze ScaffNet weights, (iii) train FusionNet with frozen ScaffNet on real data. This is denoted in row 3 of \tabref{tab:ablation_training_method_kitti_validation}. We do this consciously to enable ScaffNet to be flexible so that it can not only be used as part of the FusionNet inference pipeline, but also for other tasks that would benefit from a topology prior. Moreover, freezing ScaffNet allows for faster training with lesser hardware requirements, whereas, joint training requires extra memory, compute, and time. That being said, if we were to (i) pretrain ScaffNet on synthetic data, (ii) freeze ScaffNet weights, (iii), train FusionNet with frozen ScaffNet on real data and (iv) unfreeze ScaffNet and finetune ScaffNet and FusionNet jointly on real data, as denoted by row 4 of \tabref{tab:ablation_training_method_kitti_validation}, there can be some slight improvements to performance. However, finetuning ScaffNet jointly with FusionNet can also cause harm to the performance due to the image reconstruction term $l_{ph}$ (Eqn. 5, main text), which is susceptible to transparent, specular surfaces. Hence, the gradients computed for these terms, backpropagated through FusionNet to ScaffNet, will destroy the topology information learned from synthetic data in ScaffNet. We show this phenonmenon in \figref{fig:plot_mae_finetune_scaffnet_fusionnet}.

We further consider the the case of training ScaffNet and FusionNet end-to-end on real data to quantify the effect of using synthetic data on the full model. Quantitative results are shown in row 1 of \tabref{tab:ablation_training_method_kitti_validation}. Training ScaffNet and FusionNet jointly from scratch (without synthetic data), performs worse than training ScaffNet on synthetic data first to learn sparse geometry to dense topology (rows 2, 3, 4). This is because the model cannot leverage \textit{side information} learned from the ground-truth annotations (that come for free) in synthetic data. The results in \tabref{tab:ablation_training_method_kitti_validation} validates the proposed approach of using synthetic data to learn topology and demonstrates the benefits for using synthetic data in multi-modal sensor fusion for 3D reconstruction tasks.

\begin{table}[t]
    \caption{Results on the NYUv2 test set}
    \centering
    \setlength\tabcolsep{3.5pt}
    \begin{tabular}{l c c c c c c c}
            \midrule 
            Method & Dataset & \# Epoch & MAE & RMSE & iMAE & iRMSE \\ \midrule
            ScaffNet & SceneNet & 0 & 136.55 & 240.63 & 30.77 & 59.10 \\ \midrule
            FusionNet & VOID & 0 & 155.20 & 241.42 & 31.77 & 52.62 \\
            \midrule
            FusionNet & NYUv2 & 10 & 117.49 & 199.31 & \textbf{24.89} & \textbf{44.06} \\
            \midrule
            FusionNet & VOID, NYUv2 & 1 & 124.46 & 205.27 & 27.37 & 47.56 \\
            \midrule
            FusionNet & VOID, NYUv2 & 10 & \textbf{116.58} & \textbf{198.20} & 24.88 & 44.25 \\ 
            \midrule
        \end{tabular}
        \begin{tablenotes}
        We evaluate Scaffnet and FusionNet on the NYUv2 test set. \# Epoch denotes the number of epochs the method is trained or finetuned on NYUv2. \textbf{Row 1}: ScaffNet is trained only on SceneNet and is able to generalize again from synthetic to real dataset. \textbf{Row 2}: FusionNet is trained on VOID using a frozen ScaffNet (from row 1). \textbf{Row 3}: FusionNet is trained on NYUv2 using a frozen ScaffNet trained on SceneNet. \textbf{Row 4, 5}: FusionNet is trained on VOID using a frozen ScaffNet  (from row 1) and finetuned on NYUv2 for 1 and 10 epochs, respectively. VOID and NYUv2 are comprised of different distribution of scenes and are captured with different camera and depth sensor. Hence, while FusionNet trained on VOID (row 2) performs reasonably, it is worse than ScaffNet trained on SceneNet (row 1) due to the photometric domain gap. However, FusionNet trained on VOID can quickly adapt to the new cameras and scenes when finetuned on NYUv2 for just 1 epoch (row 4). When finetuned for 10 epochs on NYUv2 (row 5), FusionNet trained on VOID performs comparably to FusionNet trained on NYUv2.
    \end{tablenotes}
\label{tab:nyuv2_test_set}
\vspace{-1em}
\end{table}

\begin{figure*}[ht]
    \centering
    \includegraphics[height=0.3\textheight,width=1.00\linewidth]{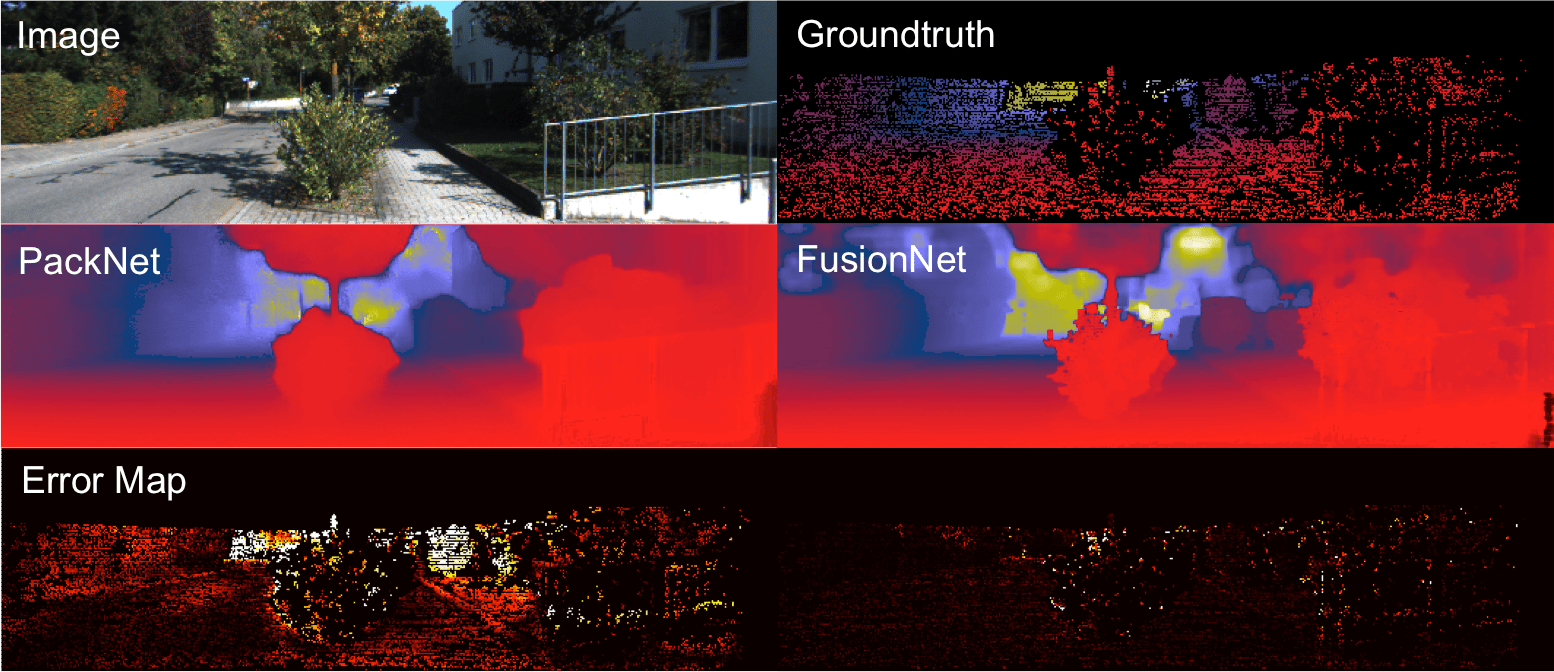}
    \caption{\textit{Qualitative comparison to state-of-the-art monocular depth network.} The input images are shown in the first (top) row, predictions of PackNet \cite{guizilini20203d} (left) and our FusionNet (right) are shown in the second (middle) row and the corresponding error maps are given in the third (bottom) row. Our method is able to recover sharper depth maps i.e. the bush in the center of the image, and vegetation and railings on the right side of the image. We also perform better overall as seen in the error maps.}
\label{fig:comparison_to_mono_networks}
\end{figure*}

\section{Generalization of ScaffNet and FusionNet to Difference Scenes}
\label{sec:generalization_studies}
In this section, we consider the generalization capabilities of ScaffNet and FusionNet for different scene distributions captured by different camera and depth sensor set ups. To examine such, we consider VOID \cite{wong2020unsupervised} and NYUv2 \cite{silberman2012indoor} for the depth completion task. VOID is comprised of scenes found in an university campus, including, but not limited to classrooms, laboratories, copy rooms, staircases, gardens, and courtyards; whereas, NYUv2 consists of household and commercial areas such as bedrooms, living rooms, dental offices, and stores. VOID is captured using an Intel RealSense D435i and NYUv2 is captured with a Microsoft Kinect. In addition to the differences in the scene distributions, and camera and depth sensor set ups, NYUv2 images also have white borders, and VOID does not. Hence, there exists a domain gap (mainly photometric) between the two datasets. The goal of this section is to understand the effects of this domain gap on performance, and whether it is possible to adapt (not just to the scenes, but also to new equipment) and recover performance.

\section{Comparisons with Monocular Depth Prediction}
\label{sec:comparison_monocular}
Here, We compare quantitatively (\tabref{tab:comparison_to_mono_networks}) and qualitatively (\figref{fig:comparison_to_mono_networks}) to the state-of-the-art monocular (single image) depth prediction models, PackNet \cite{guizilini20203d} and Monodepth2 \cite{godard2019digging}, on the KITTI validation set. For this comparison, we use the PackNet and Monodepth2 models, pretrained on KITTI, provided by the authors. 

We note that there are some differences in evaluation protocols between those typically used in monocular depth prediction literature and the ones that we employ. For instance, the metrics used here are in millimeters; whereas, those used in monocular depth prediction literature are in meters. Also, because there is a scale ambiguity in the video-based monocular depth models, it is common to perform scale matching between the predictions and ground truth. In \tabref{tab:comparison_to_mono_networks}, we do not perform scale matching because (i) Monodepth2 is trained using stereo pairs, where scale can be directly learned, and (ii) while PackNet is trained on video sequences, it also uses inertials where scale is observable. Thus, both models, like ours, should output predictions in metric scale and so we forgo the use of scale matching for fair comparison. 

In \tabref{tab:comparison_to_mono_networks}, we evaluate Monodepth2 and PackNet on the KITTI depth completion validation set using standard protocol described in the main paper. While scale can be learned through stereo training, Monodepth2 performs poorly because the scale of its predictions is off. PackNet uses velocity to learn a prior on scale so predictions are closer to metric scale, but performance is still 6 times worse than ScaffNet in MAE. This is surprising because ScaffNet does not conditioned on the ``dense'' image at all, but only the sparse point cloud. The best performing model is FusionNet, which is conditioned on both image and sparse depth. 

\figref{fig:comparison_to_mono_networks} shows a qualitative head-to-head comparison between PackNet and our method. As we can see, PackNet does recover the general shape of the scene, but structures are often over-smoothed (see bush in the center of the image, and vegetation and railing located in right side of the image). Our method produces depth maps with higher detail that capture the leaves and branches of the plants and the thin structure of the railings. The error maps (row 3) shows that our method performs better overall.

\begin{table}[t]
    \caption{Comparison with monocular depth methods on KITTI}
    \centering
    \setlength\tabcolsep{8.5pt}
    \begin{tabular}{l c c c c c}
        \midrule 
        Method & MAE & RMSE & iMAE & iRMSE \\ 
        \midrule
        Mono2 \cite{godard2019digging} & 15491.00 & 19151.41 & 8844.36 & 9890.20 \\ 
        \midrule 
        PackNet \cite{guizilini20203d} & 1808.63 & 3883.92 & 6.93 & 10.20 \\ 
        \midrule 
        Our ScaffNet & 318.41 & 1425.53 & 1.39 & 5.01 \\ 
        \midrule
        Our FusionNet & \textbf{286.35} & \textbf{1182.81} & \textbf{1.18} & \textbf{3.55} \\ 
        \midrule
    \end{tabular}
    \begin{tablenotes}
        We compare the proposed depth completion methods Scaffnet and FusionNet with state-of-the-art monocular (single image only) depth networks on the KITTI validation set. Since the scale is off for Monodepth2 \cite{godard2019digging}, it performs poorly and MAE is $\approx 15K$. PackNet \cite{guizilini20203d} performs relatively well, but still 6 times worse than Scaffnet (only using sparse depth e.g. lidar measurements) in MAE. The best performing method is FusionNet, which uses both image and sparse depth. 
    \end{tablenotes}
    \label{tab:comparison_to_mono_networks}
\end{table}

\begin{figure*}[th]
    \centering
    \includegraphics[width=1\linewidth]{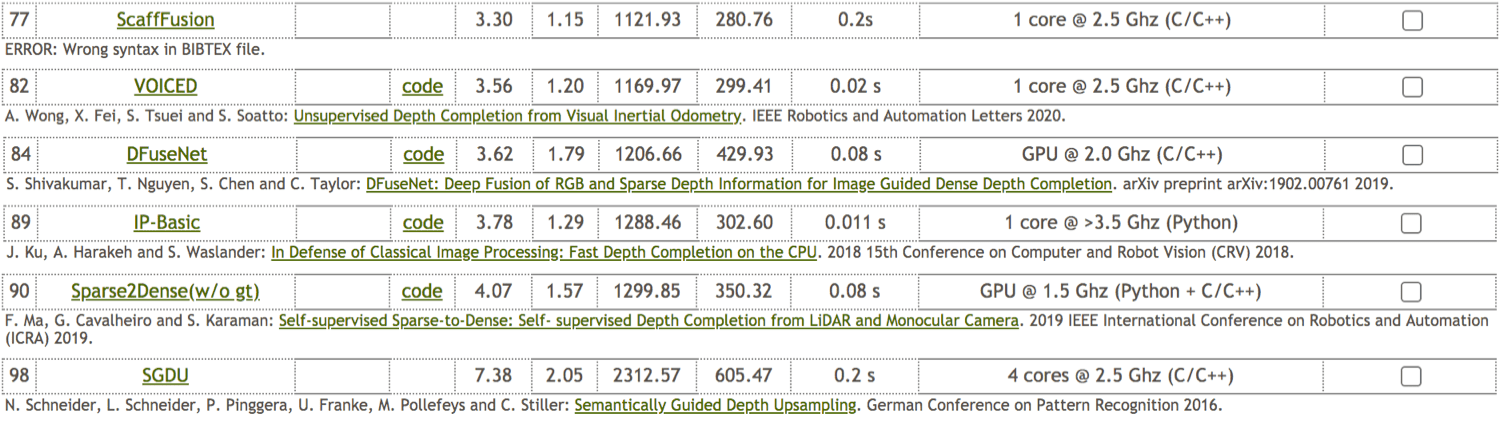}
    \caption{\textit{Compilation of Screenshots of the KITTI online benchmark at the time of submission.} Our method (ScaffFusion) ranks higher than all competing unsupervised methods. The figure was produced by concatenating screenshots of published unsupervised depth completion method on the benchmark.}
    \label{fig:kitti_benchmark_screenshot}
\end{figure*}

\begin{table}[t]
    \caption{Supervised KITTI Depth Completion Benchmark}
    \centering
    \setlength\tabcolsep{9pt}
    \begin{tabular}{l c c c c}
        \midrule 
        Method & MAE & RMSE & iMAE & iRMSE \\ 
        \midrule
        Chodosh \cite{chodosh18deep}
        & 439.48 & 1325.37 & 3.19 & 59.39 \\ 
        \midrule
        Dimitrievski \cite{dimitrievski2018learning}
        & 310.49 & 1045.45 & 1.57 & 3.84 \\ 
        \midrule 
        \textit{Our FusionNet}
        & \textit{280.76} & \textit{1121.93} & \textit{1.15}  & \textit{3.30} \\
        \midrule
        Ma \cite{ma2019self} 
        & 249.95 & 814.73 & 1.21 & 2.80	\\ 
        \midrule
        Qui \cite{qiu2019deeplidar}
        & 226.50 & 758.38 & 1.15 & 2.56	 \\ 
        \midrule
        Xu \cite{xu2019depth}
        & 235.73 & 785.57 & 1.07 & 2.52 \\ 
        \midrule			
        Chen \cite{chen2019learning}   
        & 221.19 & 752.88 & 1.14 & 2.34 \\ 
        \midrule
        Van Gansbeke \cite{van2019sparse}
        & 215.02 & 772.87 & 0.93 & 2.19	\\ 
        \midrule
        Yang \cite{yang2019dense}
        & \textbf{203.96} & 832.94 & \textbf{0.85} & 2.10 \\ 
        \midrule		
        Cheng \cite{cheng2019cspn++} 
        & 209.28 & \textbf{743.69} & 0.90 & \textbf{2.07} \\ 
        \midrule
    \end{tabular}
    \begin{tablenotes}
        \textit{Quantitative results on the \textbf{supervised} KITTI depth completion benchmark}. All results are taken from the online benchmark \cite{uhrig2017sparsity}. Methods are ordered based on all metrics rather than just RMSE (ordering of benchmark). We note \cite{ma2019self,yang2019dense} compete in both unsupervised and supervised benchmarks. We compare our \textit{unsupervised} method (italized, row 3 in the table) against supervised methods on the KITTI depth completion benchmark. We also note that while most supervised methods still do better, our approach surpasses some \textit{supervised} methods: \cite{chodosh18deep} across all metrics, \cite{dimitrievski2018learning} on MAE, iMAE, and iRMSE metrics and \cite{ma2019self} on iMAE. This demonstrates the potential of our method in closing the gap between supervised and unsupervised methods.
    \end{tablenotes}
\label{tab:kitti_supervised_benchmark_results}
\end{table}

\begin{table}[t]
    \caption{Unsupervised KITTI Depth Completion Benchmark}
    \centering
    \setlength\tabcolsep{10pt}
    \begin{tabular}{l c c c c}
        \midrule
        Method & MAE & RMSE & iMAE & iRMSE \\ 
        \midrule
        Schneider \cite{schneider2016semantically}   
        & 605.47 & 2312.57 & 2.05 & 7.38 \\ 
        \midrule
        Ma \cite{ma2019self}
        & 350.32 & 1299.85 & 1.57 & 4.07 \\ 
        \midrule
        Ku \cite{ku2018defense} 
        & 302.60 & 1288.46 & 1.29 & 3.78 \\ 
        \midrule
        Shivakumar \cite{shivakumar2019dfusenet}
        & 429.93 & 1206.66 & 1.79 & 3.62 \\ 
        \midrule
        Yang \cite{yang2019dense}
        & 343.46 & 1263.19 & 1.32 & 3.58 \\ 
        \midrule
        VGG8 \cite{wong2020unsupervised}  
        & 304.57 & 1164.58 & 1.28 & 3.66 \\ 
        \midrule
        VGG11 \cite{wong2020unsupervised}  
        & 299.41 & 1169.97 & 1.20 & 3.56 \\ 
        \midrule
        Our FusionNet
        & \textbf{280.76} & \textbf{1121.93} & \textbf{1.15}  & \textbf{3.30} \\
        \midrule
    \end{tabular}
    \begin{tablenotes}
        \textit{Quantitative results on the \textbf{unsupervised} KITTI depth completion benchmark}. Results are taken from the benchmark \cite{uhrig2017sparsity}. Our approach beats all competing methods across all metrics and achieves the state of the art on the unsupervised depth completion task. We note \cite{ma2019self,yang2019dense} compete in both unsupervised and supervised benchmarks and that the unsupervised approach of \cite{yang2019dense} also uses synthetic data. However, as \cite{yang2019dense} used both image and sparse depth from the synthetic domain, they were plagued by the domain gap between synthetic and real data. We outperform them on all metrics while using fewer parameters (also see Table II and III in main text) and we attribute some of such successes to the way we leveraged synthetic data -- bypassing the need to adapt to different domains and learning to predict topology from only sparse points.
    \end{tablenotes}
\label{tab:kitti_unsupervised_benchmark_results}
\end{table}

\section{KITTI Depth Completion Benchmark}
\label{sec:unsupervised_kitti_benchmark}

\begin{figure*}[t]
    \centering
    \includegraphics[width=0.95\linewidth]{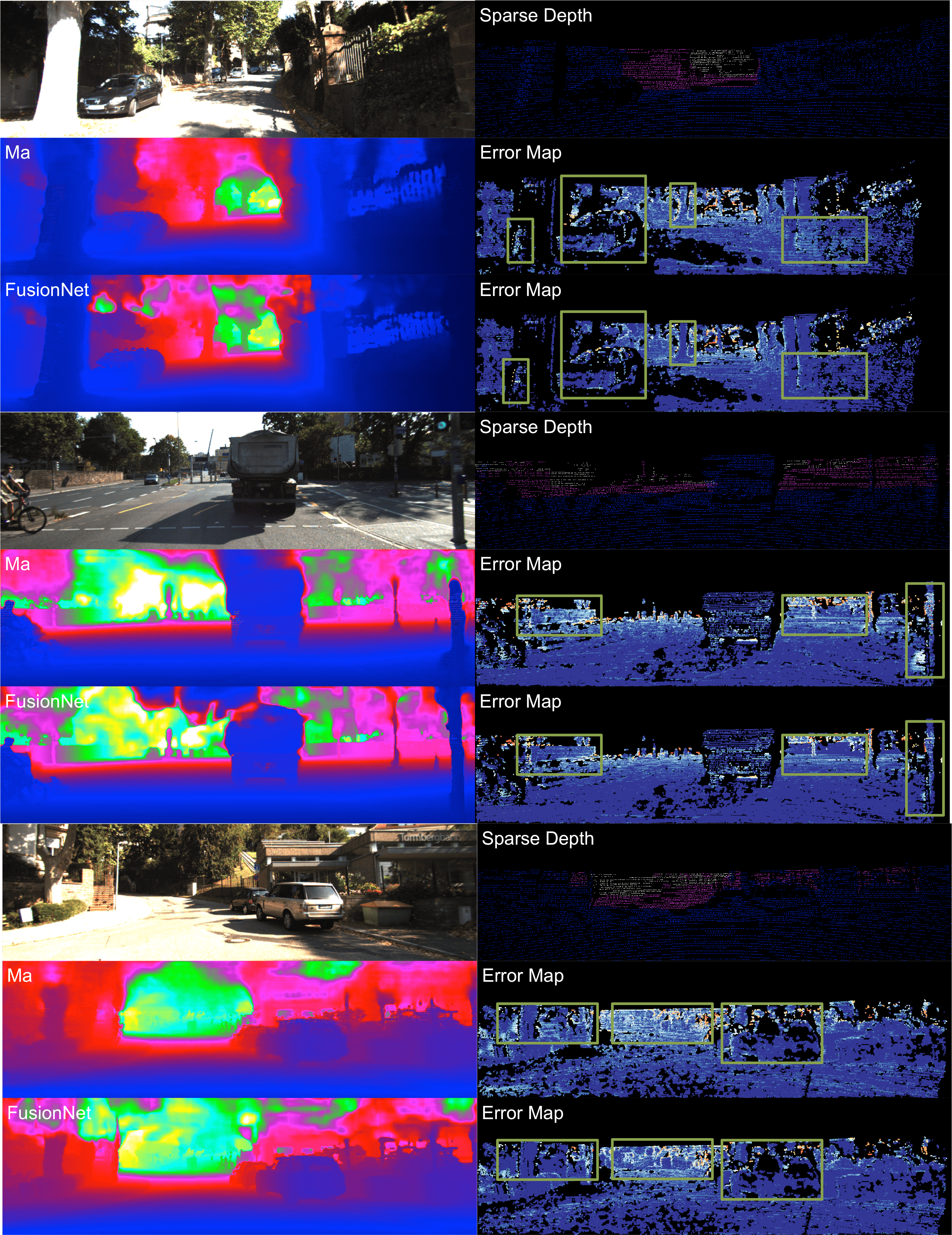}
    \caption{\textit{Qualitative comparison with \cite{ma2019self}} (best viewed $2\times$ and in color). Our approach performs better on cars, walls, trees, poles, and far regions (highlighted in green on error maps). Specifically, we show lower errors (dark blue) on the cars and walls than \cite{ma2019self} (white and light blue). In regions of the error map corresponding to walls and poles, we similarly improve (white, light blue, dark blue) over \cite{ma2019self} (red, orange, white).}
    \label{fig:kitti_testing_results_ma}
    \vspace{-1em}
\end{figure*}

\begin{figure*}[t]
    \centering
    \includegraphics[width=0.95\linewidth]{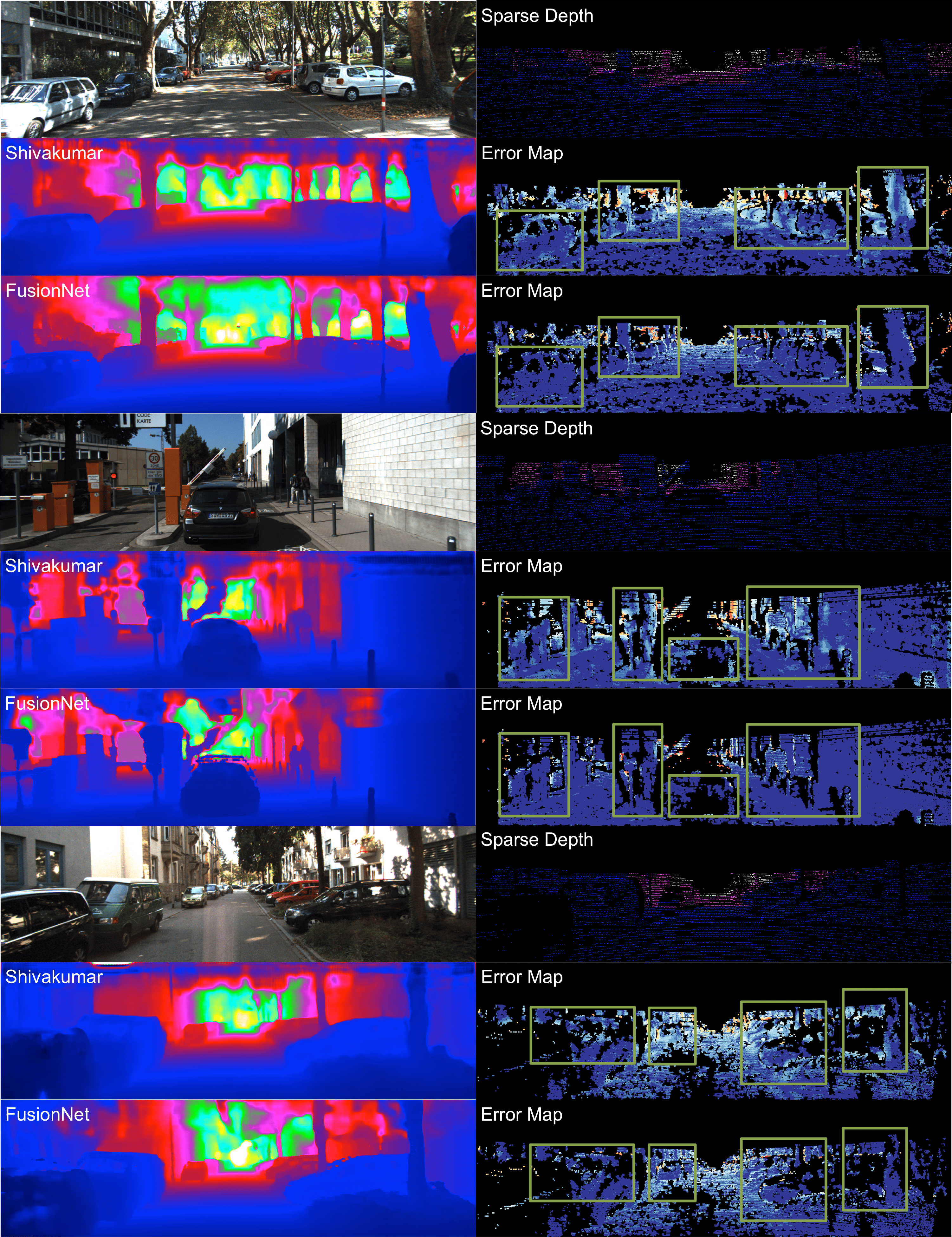}
    \caption{\textit{Qualitative comparison with \cite{shivakumar2019dfusenet}} (best viewed $2\times$ and in color). Our approach performs better on cars, walls, trees, poles, and far regions (highlighted in green on error maps). In top 3 rows, \cite{shivakumar2019dfusenet} shows errors (white) in the trees and cars. In rows 4 to 7, \cite{shivakumar2019dfusenet} shows similar levels of errors in regions corresponding to the car, walls and traffic gate. Similar errors are present in the cars in the last 3 rows. Our approach consistently performs better (dark blue) for the corresponding regions in the error maps.}
    \label{fig:kitti_testing_results_shivakumar}
    \vspace{-1em}
\end{figure*}

We showed quantitative comparisons against state-of-the-art unsupervised methods in Table II and III of the main text. Here, we compare our approach against the both top performing \textit{supervised} (\tabref{tab:kitti_supervised_benchmark_results}) and \textit{unsupervised} (\tabref{tab:kitti_unsupervised_benchmark_results}) methods on the KITTI depth completion benchmark \cite{uhrig2017sparsity}. Our method is the state of the art on the unsupervised depth completion task, outperforming all competing methods across all metrics. We note \cite{ma2019self,yang2019dense} compete in both unsupervised and supervised benchmarks. A key comparison is the unsupervised approach of \cite{yang2019dense}, who also used synthetic data (both image and depth), yet, we outperform them across all metrics. We attribute our successes over their method to the way we exploited synthetic data -- learning to recover topology from sparse points. As \cite{yang2019dense} used both image and sparse depth from the synthetic domain, they were plagued by the domain gap between synthetic and real data. Our approach, instead, leverages geometry from the synthetic domain, where shapes of objects persist regardless of real or synthetic domains. Hence, our model is able to bridge the domain gap without any sort of adaptation. Although we are the best performing method on the unsupervised setting, we note that the benchmark is still dominated by supervised methods. However, our approach shows promise as we surpass some supervised methods: \cite{chodosh18deep} across all metrics, \cite{dimitrievski2018learning} on MAE, iMAE, and iRMSE metrics and \cite{ma2019self} on iMAE. We hope that our approach will lay the foundation for the push to close the gap between supervised and unsupervised learning frameworks for the depth completion task.

In \figref{fig:kitti_benchmark_screenshot}, we compiled screenshots of each of the competing unsupervised depth completion method and their respective ranks on the KITTI benchmark at the time of submission. Individual screenshots of each were concatenated together to form the figure. The unsupervised version of \cite{yang2019dense} is omitted because they did not release their results on the online benchmark. Their numbers were directly taken from their paper and are shown in \tabref{tab:kitti_unsupervised_benchmark_results}. Our method (ScaffFusion) is the state of the art on the unsupervised depth completion task.

In the main paper, we showed qualitative comparisons with the state of the art, VGG11 \cite{wong2020unsupervised}. Here, in \figref{fig:kitti_testing_results_ma} and \ref{fig:kitti_testing_results_shivakumar}, we show additional comparisons with other top unsupervised depth completion methods \cite{ma2019self,shivakumar2019dfusenet} on the KITTI online benchmark \cite{uhrig2017sparsity}. Compared to \cite{ma2019self}, our approach performs better on cars, walls, trees, poles, and far regions. Also, in the predicted depth maps, \cite{ma2019self} shows artifacts resembling scanlines. Compared to \cite{shivakumar2019dfusenet}, we similarly improve on cars, walls, trees and poles. \cite{shivakumar2019dfusenet} tend to predict irregular shapes incorrectly, as seen in the error map of the traffic gate in rows 4 to 7 in \figref{fig:kitti_testing_results_shivakumar}; whereas, we capture its geometry. Most of these regions are complex and thus demonstrates the effectiveness of our topology estimator, ScaffNet, (Sec. III-A in main text) and the selective use our topology prior (Eqn. 8, 9 in main text) in our objective function. Error maps in \figref{fig:kitti_testing_results_ma} and \ref{fig:kitti_testing_results_shivakumar} have been highlighted in green to show the comparisons. 

\section{Network Architecture}
\label{sec:network_architectures}
We present our network architectures for our topology estimator, and our depth-RGB image fusion network, ScaffNet and FusionNet, respectively (see Fig. II in main text). We design our network structures carefully to make them light-weight so that they can be employed on standard embedded systems for real-time applications. 

Our ScaffNet is an SPP module followed by an encoder and decoder, which all together consists of only $\approx$1.4M parameters. Given sparse depth measurements, our ScaffNet outputs an initial estimate of the topology, $\hat{d}_{0}$, which is refined by our FusionNet. Our FusionNet follows the late fusion paradigm with an image branch and a depth branch. The latent representation of both branches are concatenated and fed to decoder. The decoder produces multiplicative scales $\alpha(x)$ and additive residuals $\beta(x)$ to construct our final depth prediction  $\hat{d}(x) = \alpha(x) \hat{d}_{0}(x) + \beta(x)$ for $x \in \Omega$. Our FusionNet is extremely light-weight and only consists of $\approx$6.4M parameters. Together, our model has a total of $\approx$7.8M parameters, much fewer than the competing methods $\approx$9.7M \cite{wong2020unsupervised}, $\approx$18.8M \cite{yang2019dense}, and $\approx$27.8M \cite{ma2019self} while outperforming all of them (\tabref{tab:kitti_unsupervised_benchmark_results})) across all metrics.

\subsection{ScaffNet} 
ScaffNet is an encoder-decoder network augmented with an Spatial Pyramid Pooling module. Our SPP module is comprised of several max pool layers followed by three $1 \times 1$ convolutional layers to weight the trade-off between small (fine details, but sparse) and large (coarse, but dense) kernel sizes. The output of our SPP module is fed into an encoder with five layers. The first layer uses a $5 \times 5$ kernel with 32 filters. The remaining four layers use $3 \times 3$ kernels with 64, 96, 128, and 196 filters, respectively. The latent representation is then fed into a decoder with skip connections. The decoder is separated into five modules, each corresponds to a resolution level. For each module, we perform de-convolution using $3 \times 3$ kernels with 128, 96, 64, 64, and 32 filters, respectively. The result is concatenated with the output of the corresponding resolution (skip connection) from the encoder. The output of which is convolved with another $3 \times 3$ and passed to the next decoder module.

\subsection{FusionNet}
Our FusionNet consists of two encoders (branches), one for processing image and the other for depth. Both encoders contain five convolutional layers with the first layer using a $5 \times 5$ kernel. The remaining four layers for each branch use $3 \times 3$ kernels. For the image encoder, the convolutional layers consist of 48, 96, 192, 384, and 384 filters. For the depth encoder, the layers consist of 16, 32, 64, 128, and 128 filters. The output latent representations of the encoders are concatenated together and fed into the decoder. The FusionNet decoder, unlike the ScaffNet decoder, contains four modules, where each of the modules is comprised of $3 \times 3$ de-convolution, followed by concatenation with skip connections, and a $3 \times 3$ convolution. The skip connections are the concatenation of the feature maps from the corresponding image and depth encoders. The de-convolution and convolution layers of each module use 256, 128, 64, 64 filters, respectively. The low resolution output is produced by a $3 \times 3$ convolution with 2 filters, for $\alpha$ and $\beta$. The last layer is a nearest-neighbor upsampling layer to produce the full resolution $\alpha$ and $\beta$.  

\begin{table*}[h!]
    \centering
    \normalsize
    \setlength\tabcolsep{12pt}
    \begin{tabular}{lcccccccc@{}}
        \textbf{ScaffNet} & \multicolumn{2}{c}{kernel} & \multicolumn{2}{c}{channels} & \multicolumn{2}{c}{resolution} & \\ 
        \cmidrule(lr){2-3} 
        \cmidrule(lr){4-5}
        \cmidrule(lr){6-7}
        layer & size & stride & in & out & in & out & \# params & input \\ 
        \midrule
        \textit{SPP Module} & \multicolumn{1}{l}{} & \multicolumn{1}{l}{} & \multicolumn{1}{l}{} & \multicolumn{1}{l}{} & \multicolumn{1}{l}{} & \multicolumn{1}{l}{} & \multicolumn{1}{l}{} \\
        \midrule
        pool1        & 5  & 1 & 2         & 2   & 1    & 1 & 0             & depth \\ \midrule
        pool2        & 7  & 1 & 2         & 2   & 1    & 1 & 0             & depth \\ \midrule
        pool3        & 9  & 1 & 2         & 2   & 1    & 1 & 0             & depth \\ \midrule
        pool4        & 11 & 1 & 2         & 2   & 1    & 1 & 0             & depth \\ \midrule
        concat       & -  & - & 2+2+2+2+2 & 10  & \multicolumn{4}{c}{depth, pool1, pool2, pool3, pool4} \\ \midrule 
        conv1        & 1 & 1  & 10        & 32  & 1    & 1 & 320           & concat  \\ \midrule
        conv2        & 1 & 1  & 32        & 32  & 1    & 1 & $\approx$ 1K  & conv1  \\ \midrule
        output\_ssp  & 1 & 1  & 32        & 32  & 1    & 1 & $\approx$ 1K  & conv2  \\ \midrule
        \textit{Encoder} & \multicolumn{1}{l}{} & \multicolumn{1}{l}{} & \multicolumn{1}{l}{} & \multicolumn{1}{l}{} & \multicolumn{1}{l}{} & \multicolumn{1}{l}{} & \multicolumn{1}{l}{} \\
        \midrule
        conv1  & 5 & 2 & 32  & 32  & 1    & 1/2  & $\approx$ 25K    & output\_ssp  \\ \midrule
        conv2  & 3 & 2 & 32  & 64  & 1/2  & 1/4  & $\approx$ 18K    & conv1  \\ \midrule
        conv3  & 3 & 2 & 64  & 96  & 1/4  & 1/8  & $\approx$ 55K    & conv2  \\ \midrule
        conv4  & 3 & 2 & 96  & 128 & 1/16 & 1/16 & $\approx$ 110K   & conv3  \\ \midrule
        latent & 3 & 2 & 128 & 196 & 1/16 & 1/32 & $\approx$ 225K   & conv4  \\ \midrule
        \textit{Decoder} & \multicolumn{1}{l}{} & \multicolumn{1}{l}{} & \multicolumn{1}{l}{} & \multicolumn{1}{l}{} & \multicolumn{1}{l}{} & \multicolumn{1}{l}{} & \multicolumn{1}{l}{} \\
        \midrule
        deconv5 & 3 & 2 & 196         & 128 & 1/32 & 1/16  & $\approx$ 225K & latent   \\ \midrule
        concat5 & - & - & 128+128     & 256 & \multicolumn{4}{c}{deconv5, conv4}       \\ \midrule
        conv5   & 3 & 1 & 256         & 128 & 1/16 & 1/16  & $\approx$ 250K & concat5  \\ \midrule
        deconv4 & 3 & 2 & 128         & 96  & 1/16 & 1/8   & $\approx$ 110K & conv5    \\ \midrule
        concat4 & - & - & 96+96       & 192 & \multicolumn{4}{c}{deconv4, conv3}       \\ \midrule
        conv4   & 3 & 1 & 192         & 96  & 1/8  & 1/8   & $\approx$ 166K & concat4  \\ \midrule
        deconv3 & 3 & 2 & 96          & 64  & 1/8  & 1/4   & $\approx$ 55K  & conv4    \\ \midrule
        concat3 & - & - & 64+64       & 128 & \multicolumn{4}{c}{deconv3, conv2}       \\ \midrule
        conv3   & 3 & 1 & 128         & 64  & 1/4  & 1/4   & $\approx$ 78K  & concat3  \\ \midrule
        deconv2 & 3 & 2 & 64          & 64  & 1/4  & 1/2   & $\approx$ 37K  & conv3    \\ \midrule
        concat2 & - & - & 64+32       & 96  & \multicolumn{4}{c}{deconv2, conv1}       \\ \midrule
        conv2   & 3 & 1 & 96          & 64  & 1/2  & 1/2   & $\approx$ 55K  & concat2  \\ \midrule
        deconv1 & 3 & 2 & 64          & 32  & 1/2  & 1     & $\approx$ 18K  & conv2    \\ \midrule
        output  & 3 & 1 & 32          & 32  & 1    & 1     & $\approx$ 9K   & deconv1  \\
        \midrule 
        \midrule
        Total Parameters & $\approx$ 1.4M
    \end{tabular}
    \begin{tablenotes}
        We note that the max pooling layers specified here ($5 \times 5$, $7 \times 7$, $9 \times 9$, and $11 \times 11$) are designed for KITTI \cite{uhrig2017sparsity}. We use an extra max pooling layer with kernel size $13 \times 13$ for the VOID dataset \cite{wong2020unsupervised}. 
    \end{tablenotes}
\end{table*}

\begin{table*}[h!]
    \centering
    \normalsize
    \setlength\tabcolsep{11pt}
    \begin{tabular}{lcccccccc@{}}
        \textbf{FusionNet} & \multicolumn{2}{c}{kernel} & \multicolumn{2}{c}{channels} & \multicolumn{2}{c}{resolution} & \\ 
        \cmidrule(lr){2-3} 
        \cmidrule(lr){4-5}
        \cmidrule(lr){6-7}
        layer & size & stride & in & out & in & out & \# params & input \\ 
        \midrule
        \textit{Image Encoder} & \multicolumn{1}{l}{} & \multicolumn{1}{l}{} & \multicolumn{1}{l}{} & \multicolumn{1}{l}{} & \multicolumn{1}{l}{} & \multicolumn{1}{l}{} & \multicolumn{1}{l}{} \\
        \midrule
        conv1\_image          & 5 & 2 & 3   & 48  & 1    & 1/2  & $\approx$ 3.6K & image  \\ \midrule
        conv2\_image          & 3 & 2 & 48  & 96  & 1/2  & 1/4  & $\approx$ 41K  & conv1\_image  \\ \midrule
        conv3\_image          & 3 & 2 & 96  & 192 & 1/4  & 1/8  & $\approx$ 166K & conv2\_image  \\ \midrule
        conv4\_image          & 3 & 2 & 192 & 384 & 1/8  & 1/16 & $\approx$ 663K & conv3\_image \\ \midrule
        latent\_image         & 3 & 2 & 384 & 384 & 1/16 & 1/32 & $\approx$ 1.3M & conv4\_image \\ \midrule
        \textit{Depth Encoder} & \multicolumn{1}{l}{} & \multicolumn{1}{l}{} & \multicolumn{1}{l}{} & \multicolumn{1}{l}{} & \multicolumn{1}{l}{} & \multicolumn{1}{l}{} & \multicolumn{1}{l}{} \\ 
        \midrule
        conv1\_depth          & 5 & 2 & 2   & 16  & 1    & 1/2  & $\approx$ 0.8K & depth  \\ \midrule
        conv2\_depth          & 3 & 2 & 16  & 32  & 1/2  & 1/4  & $\approx$ 4.6K & conv1\_depth  \\ \midrule
        conv3\_depth          & 3 & 1 & 32  & 64  & 1/4  & 1/4  & $\approx$ 18K  & conv2\_depth  \\ \midrule
        conv4\_depth          & 3 & 1 & 64  & 128 & 1/8  & 1/16 & $\approx$ 74K  & conv3\_depth \\ \midrule
        latent\_depth         & 3 & 2 & 128 & 128 & 1/16 & 1/32 & $\approx$ 147K & conv4\_depth \\ \midrule
        \textit{Latent Encoding} & \multicolumn{1}{l}{} & \multicolumn{1}{l}{} & \multicolumn{1}{l}{} & \multicolumn{1}{l}{} & \multicolumn{1}{l}{} & \multicolumn{1}{l}{} & \multicolumn{1}{l}{} \\ \midrule
        latent         & - & - & 384+128 & 512 & \multicolumn{4}{c}{latent\_image, latent\_depth} \\ \midrule
        \textit{Decoder} & \multicolumn{1}{l}{} & \multicolumn{1}{l}{} & \multicolumn{1}{l}{} & \multicolumn{1}{l}{} & \multicolumn{1}{l}{} & \multicolumn{1}{l}{} & \multicolumn{1}{l}{} \\ \midrule
        deconv5 & 3 & 2 & 512         & 256 & 1/32 & 1/16  & $\approx$ 1.2M & latent   \\ \midrule
        concat5 & - & - & 256+384+128 & 768 & \multicolumn{4}{c}{deconv5, conv4\_image, conv4\_depth} \\ \midrule
        conv5   & 3 & 1 & 768         & 256 & 1/16 & 1/16  & $\approx$ 1.8M & concat5  \\ \midrule
        deconv4 & 3 & 2 & 256         & 128 & 1/16 & 1/8   & $\approx$ 295K & conv5    \\ \midrule
        concat4 & - & - & 128+192+64  & 384 & \multicolumn{4}{c}{deconv4, conv3\_image, conv3\_depth} \\ \midrule
        conv4   & 3 & 1 & 384         & 128 & 1/8  & 1/8   & $\approx$ 442M & concat4  \\ \midrule
        deconv3 & 3 & 2 & 128         & 128 & 1/8  & 1/4   & $\approx$ 147K  & conv4    \\ \midrule
        concat3 & - & - & 128+96+32   & 256 & \multicolumn{4}{c}{deconv3, conv2\_image, conv2\_depth} \\ \midrule
        conv3   & 3 & 1 & 256         & 64  & 1/4  & 1/4   & $\approx$ 147K & concat3  \\ \midrule
        deconv2 & 3 & 2 & 64          & 64  & 1/4  & 1/2   & $\approx$ 37K  & conv3    \\ \midrule
        concat2 & - & - & 64+48+16    & 128 & \multicolumn{4}{c}{deconv2, conv1\_image, conv1\_depth} \\ \midrule
        conv2   & 3 & 1 & 128         & 2   & 1/2  & 1/2   & $\approx$ 2.4K & concat2  \\ \midrule
        output  & - & - & -           & -   & 1/2  & 1     & \multicolumn{2}{c}{upsample conv2} \\
        \midrule 
        \midrule
        Total Parameters & $\approx$ 6.4M
    \end{tabular}
\end{table*}

\end{appendices}

\end{document}